%% file: paper_arxiv_final.tex
\documentclass{article}

     \PassOptionsToPackage{numbers, compress}{natbib}
    \usepackage[final]{neurips_2021}

\usepackage[utf8]{inputenc} % allow utf-8 input
\usepackage[T1]{fontenc}    % use 8-bit T1 fonts
\usepackage{url}            % simple URL typesetting
\usepackage{booktabs}       % professional-quality tables
\usepackage{amsfonts}       % blackboard math symbols
\usepackage{nicefrac}       % compact symbols for 1/2, etc.
\usepackage{microtype}      % microtypography
\usepackage{xcolor}         % colors
\usepackage{makecell}

\input macros

\input labelfig

\usepackage{subfigure}
\usepackage{amsmath,amssymb,amsfonts,latexsym,verbatim,color,epsfig,psfrag}%epsf,
\usepackage{dsfont}
\usepackage{times,multirow,multicol, array}
\usepackage{algorithm,algorithmic}
\usepackage{setspace}
\usepackage{float}
\usepackage{caption}
\usepackage{adjustbox}
\usepackage{wrapfig}

\newcolumntype{L}[1]{>{\raggedright\let\newline\\\arraybackslash\hspace{0pt}}m{#1}}
\newcolumntype{C}[1]{>{\centering\let\newline\\\arraybackslash\hspace{0pt}}m{#1}}
\newcolumntype{R}[1]{>{\raggedleft\let\newline\\\arraybackslash\hspace{0pt}}m{#1}}

\definecolor{ao}{rgb}{0.0, 0.5, 0.0}

\title{A Max-Min Entropy Framework for Reinforcement Learning}

\author{%
    Seungyul Han \\%\thanks{Use footnote for providing further
    Graduate School of Artificial Intelligence\\
	UNIST\\
	Ulsan, South Korea 44919\\
	\texttt{syhan@unist.ac.kr} \\
	\And
	Youngchul Sung$^{\dagger}$ \\
	Dept. of Electrical Engineering\\
	KAIST \\
	Daejeon, South Korea 34141\\
	\texttt{ycsung@kaist.ac.kr} \\
}

\begin{document}

\maketitle

\begin{abstract}
In this paper, we propose a max-min entropy framework for reinforcement learning (RL) to overcome the limitation of the soft actor-critic (SAC) algorithm  implementing the maximum entropy RL in model-free sample-based learning. Whereas the maximum entropy RL guides learning for policies to reach states with high entropy in the future, the proposed max-min entropy framework aims to learn to visit states with low entropy and maximize the entropy of these low-entropy states to promote better exploration. For general Markov decision processes (MDPs), an efficient algorithm is constructed under the proposed max-min entropy framework based on disentanglement of exploration and exploitation. Numerical results show that the proposed algorithm yields drastic performance improvement over the current state-of-the-art RL algorithms. 
\end{abstract}

%%%%%%%%%%%%%%%%%%%%%%%%%%%%%%%%%%%%%%%%
\section{Introduction}
\label{sec:intro}
%%%%%%%%%%%%%%%%%%%%%%%%%%%%%%%%%%%%%%%%

\begin{comment}
Standard reinforcement learning (RL) aims to maximize the expected (discounted) sum of reward for a given Markov decision process (MDP) environment \citep{sutton1998reinforcement}. Deep $Q$-learning (DQN), which is one of famous standard RL algorithms, learns $Q$-function to maximize the expected return and achieves human level performance in Atari games \citep{mnih2015human}. In order to guarantee the convergence of $Q$-function, exploration is an important issue to visit diverse state-action pairs because the proof of convergence assumes that the policy visits all state-action pairs infinitely often \citep{watkins1992q}. Thus, many methods based on intrinsic-motivated reward for better exploration have been studied \citep{achiam2017surprise, barto2013intrinsic, houthooft2016vime, martin2017count} and shows a good performance in various challenging environments such as Montezuma's Revenge from improved exploration recently \citep{badia2020never,burda2018exploration}. 
\end{comment}

The maximum entropy framework has been considered in various RL domains \citep{   haarnoja2018soft,haarnoja2018soft2,levine2013guided,rawlik2013stochastic, todorov2008general, toussaint2009robot,ziebart2008maximum}.
Maximum entropy RL adds the expected policy entropy to the return objective of standard RL in order to maximize both the return and the entropy of policy distribution. Maximum entropy RL encourages the policy to choose multiple actions probabilistically and yields a significant improvement in exploration and robustness and good final performance in various control tasks \citep{eysenbach2021maximum, haarnoja2018learning, haarnoja2017rein,  han2020diversity, hazan2019provably, huang2019svqn, singh2019end}.
In particular, soft actor-critic (SAC)  implements maximum entropy RL  in an efficient iterative manner based on soft policy iteration and guarantees  convergence to the  optimal policy for  finite MDPs, yielding  significant performance improvement over various on-policy and off-policy recent RL algorithms  in many continuous control tasks. 
However, we observe that such an iterative implementation of the maximum entropy strategy of optimizing for policies that aim to reach states with high  entropy in the future does not necessarily result in the desired exploration behavior but may yield positive feedback  hindering exploration in model-free sample-based learning with function approximation.  
In order to overcome such limitations  associated with implementation of  the maximum entropy RL, we  propose   
a max-min entropy framework for RL, which aims to learn policies reaching states with low entropy and maximizing the entropy of these low-entropy states, whereas the conventional maximum entropy RL optimizes for policies that aim to visit states with high entropy and maximize the entropy of those high-entropy states for  high entropy of the entire trajectory.  We implemented the proposed max-min entropy framework into a practical iterative actor-critic algorithm based on policy iteration with disentangled exploration and exploitation. It is demonstrated  that the proposed algorithm significantly enhances  exploration capability due to the fairness across states induced by the max-min framework and yields drastic performance improvement over existing RL algorithms including maximum-entropy SAC on difficult control tasks.

\section{Related Works}
\label{sec:related}

{\bf Maximum Entropy RL}:  The maximum entropy framework has been considered in various RL domains: inverse reinforcement learning \citep{ziebart2008maximum}, stochastic optimal control \citep{rawlik2013stochastic, todorov2008general, toussaint2009robot}, guided policy search \citep{levine2013guided}, and off-policy learning \citep{haarnoja2018soft,haarnoja2018soft2}. There is a connection between value-based and policy-based RL under the policy entropy regularization \citep{nachum2017bridging}, \citep{o2016combining} combines them, and finally \citep{schulman2017equivalence} proves that  they are equivalent. Maximum entropy  RL is also related to probabilistic inference \citep{neumann2011variational, rawlik2013stochastic}. Recently, maximizing the entropy of state distribution instead of the policy distribution \citep{hazan2019provably} and  maximizing the entropy considering the previous sample action distribution \citep{han2020diversity} have been investigated  for better exploration.  
\begin{comment}
Maximum entropy RL  shows good performance in various environments \citep{haarnoja2018soft, han2020diversity, singh2019end}.
\end{comment}

{\bf Max-Min Optimization}: Max-min optimization 
aims to maximize the minimum   of the objective function  \citep{chinchuluun2008pareto}. 
\begin{comment}
It is known that a saddle value exists when the variable spaces are compact convex in $\mathbb{R}^n$ \citep{kneser1952theoreme,neumann1928theorie}.\end{comment}
Under the convex-concave assumption, there exist  many algorithms to find the solution to a max-min problem 
\begin{comment}
%(Nash equilibria \citep{nash1950equilibrium} in terms of game theory)
\end{comment}
by using optimistic mirror descent \citep{rakhlin2013optimization}, Frank-Wolfe algorithm \citep{gidel2017frank}, and Primal-Dual method \citep{hamedani2018iteration}. However, non-convex max-min problems  are more challenging \citep{murty1985some} and  there are several recent studies to find  (approximate) solutions to  non-convex max-min optimization problems \citep{barazandeh2020solving, nouiehed2019solving, rafique2018non}. 
This framework has been used in various  optimization/control domains: fair resource allocation \citep{liu2013max}, inference \citep{baharlouei2019r,zhang2018mitigating}, generative adversarial network (GAN) \citep{arjovsky2017wasserstein,goodfellow2014generative}, robust training \citep{madry2017towards}, and reinforcement learning \citep{wai2019variance}.

{\bf Exploration in RL}: Exploration is one of the most important issues in model-free RL, as there is the key assumption that all state-action pairs must be visited infinitely often  to guarantee the convergence of $Q$-function \citep{watkins1992q}. In order to explore diverse state-action pairs in the joint state-action space, various methods have been considered in prior works: intrinsically-motivated reward based on curiosity \citep{baldassarre2013intrinsically, chentanez2005intrinsically}, model prediction error \citep{achiam2017surprise, burda2018exploration}, information gain \citep{han2020diversity, hong2018diversity, houthooft2016vime}, and counting states \citep{lopes2012exploration,martin2017count}. These  exploration techniques improve exploration and performance  in challenging sparse-reward environments  \citep{badia2020never, burda2018exploration, choi2018contingency}.

\section{Background}
\label{sec:background}

\subsection{Basic RL Setup}

\begin{comment}
We consider a basic RL setup that aims to maximize the discounted return in a given environment by learning an RL agent. 
\end{comment}
We consider an infinite-horizon MDP $(\mathcal{S},\mathcal{A},P,\gamma,r)$, where $\mathcal{S}$ is the state space, $\mathcal{A}$ is the action space, $P$ is the transition probability, $\gamma$ is the discount factor, and $r$ is the bounded reward function. We assume that each action dimension is bounded. The RL agent has a policy $\pi:\mathcal{S}\times\mathcal{A}\rightarrow \mathbb{R}^+ \in \Pi$, which chooses an action $a_t$ for given state $s_t$ according to $a_t \sim \pi(\cdot|s_t)$  at each time step $t$, where $\Pi$ is the policy space.  For  action $a_t$,   the environment yields  the reward $r_t:=r(s_t,a_t)$ and the next state  $s_{t+1} \sim P(s_{t+1}|s_t,a_t)$. Standard RL learns  policy $\pi$ to maximize the discounted return $\mathbb{E}_{s_0\sim p_0,~\tau_0\sim\pi}[\sum_{t=0}^\infty\gamma^t r_t]$, where $\tau_t=(s_t,a_t,s_{t+1},a_{t+1},\cdots)$ is an episode trajectory.

\subsection{Maximum Entropy RL and Soft Actor-Critic}
\label{subsec:sac}

Maximum entropy RL maximizes both the expected return and  the expected policy entropy simultaneously to achieve an improvement in exploration and robustness.
\begin{comment}
It allows the policy to select multiple actions probabilistically to explore a wider range in the action space, and can yield better performance as compared to standard RL in various environments \citep{haarnoja2017rein, haarnoja2018soft} and be robust to perturbations \citep{haarnoja2018learning, huang2019svqn}.
\end{comment}
 The entropy-augmented objective function of maximum entropy RL is given by
\begin{equation}
J_{MaxEnt}(\pi) = \mathbb{E}_{s_0\sim p_0, ~\tau_0\sim\pi}\left[\sum_{t=0}^\infty \gamma^t (r_t + \alpha \mathcal{H}(\pi(\cdot|s_t)))\right],
\label{eq:objsac}
\end{equation}
where $\mathcal{H}(\pi(\cdot|s))=\mathbb{E}_{a \sim\pi(\cdot|s)}[- \log \pi(a|s)]$ is the entropy function and $\alpha>0$ is the entropy coefficient. A key point here is that the policy entropy is included in the reward not used as an external regularizer at each time step. Thus, {\em this maximum entropy RL framework optimizes for policies that aim to reach states on which policies have high entropy in the future} \citep{haarnoja2017rein}.

Soft actor-critic (SAC)  is an  efficient   off-policy actor-critic  algorithm to solve the maximum entropy RL problem  \citep{haarnoja2018soft}. SAC  maximizes \eqref{eq:objsac} based on soft policy iteration, which consists of soft policy evaluation and soft policy improvement. For this, the soft $Q$-value of given $(s_t,a_t)$ is defined as
\begin{equation}
Q^\pi(s_t,a_t) := r_t +  \mathbb{E}_{\tau_{t+1}\sim\pi}\left[\sum_{l=t+1}^\infty\gamma^{l-t} (r_l + \alpha \mathcal{H}(\pi(\cdot|s_l)))\right],
\label{eq:softq}
\end{equation}
which does not include the policy entropy of the current time step but includes the sum of all future policy entropy and the sum of all current and future rewards.
For given $\pi$, soft policy evaluation guarantees the convergence of  soft $Q$-function estimation, which estimates $Q^\pi$ by iteratively applying a modified Bellman  operator $\mathcal{T}^\pi$ to a real-valued estimate function $Q:\mathcal{S}\times\mathcal{A}\rightarrow \mathbb{R}$, given by  
\begin{align}
    \mathcal{T}^\pi Q(s_t,a_t) &= r_t + \gamma \mathbb{E}_{s_{t+1}\sim P(\cdot|s_t,a_t)}[ V(s_{t+1})],~~~~~\mbox{where}\label{eq:bellman}\\
    V(s_t) &= \mathbb{E}_{a_t\sim\pi(\cdot|s_t)}[Q(s_t,a_t)-\alpha\log\pi(a_t|s_t)] \nonumber
\end{align}
and $V(s_t)$ is the soft state value function.  Then,  at each iteration, SAC updates the policy  as
\begin{align}
    \pi_{new} &= \mathop{\arg\min}_{\pi\in\Pi}D_{KL} \left(\pi(\cdot|s_t) || \frac{\exp(Q^{\pi_{old}}(s_t,a_t)/\alpha)}{Z^{\pi_{old}}(s_t)}  \right)\\
    &=\mathop{\arg\max}_{\pi\in\Pi}\mathbb{E}_{a_t\sim \pi(\cdot|s_t)}[Q^{\pi_{old}}(s_t,a_t)-\alpha\log\pi(a_t|s_t)]
\label{eq:sacpolup}
\end{align}
where $Z^{\pi_{old}}(s_t)$ is the log partition function which is a function of  $s_t$ only.
Soft policy improvement guarantees $Q^{\pi_{new}}(s_t,a_t)\geq Q^{\pi_{old}}(s_t,a_t)$ for all $(s_t,a_t)\in\mathcal{S}\times\mathcal{A}$.
Finally,  soft policy evaluation and soft policy improvement are repeated. Then, any initial policy $\pi \in \Pi$ converges to the optimal policy $\pi^*$, i.e., $Q^{\pi^{*}}(s_t,a_t)\geq Q^{\pi'}(s_t,a_t)$ for all $\pi'\in\Pi$ and all $(s_t,a_t)\in \mathcal{S}\times\mathcal{A}$, and  $\pi^*$ maximizes $J_{MaxEnt}$ \citep{haarnoja2018soft}. Proof of soft policy iteration  assumes finite MDPs.  SAC approximates the soft policy iteration by sample-based learning with function approximation  in continuous-space cases.  
\begin{comment}
SAC stores $(s_t,a_t,r_t,s_{t+1})$ in a replay buffer $\mathcal{D}$, and samples a mini-batch from the buffer for $Q$-value estimation in \eqref{eq:bellman} and the policy update in \eqref{eq:sacpolup} at each time step $t$. For the policy update, the estimated $Q$-function  replaces $Q^{\pi_{old}}$ in \eqref{eq:sacpolup}.
\end{comment}

\section{Motivation: Limitation of Maximum Entropy SAC in Pure Exploration}
\label{sec:motivation}

In this section, we will consider only the maximum entropy SAC in a pure exploration setup without the reward function (the reward function $r=0$ in MDPs). As seen in Sec. \ref{sec:background}, 
SAC efficiently solves the maximum entropy RL problem to maximize \eqref{eq:objsac} in an iterative manner based on judiciously-defined state and action value functions and the step-wise optimization cost \eqref{eq:sacpolup}. 
 The well-defined value functions and the local cost function as such enable proof of soft policy improvement for finite MDPs in a similar way to the proof of the classical policy improvement theorem. Note that at each time step, SAC updates the policy to maximize the cost  \eqref{eq:sacpolup}, composed of two terms: $\mathbb{E}_{a_t \sim \pi(\cdot|s_t)} [Q^{\pi_{old}}(s_t,a_t)]$ and
 $\alpha \mathbb{E}_{a_t \sim \pi(\cdot|s_t)}[-\log \pi (a_t|s_t)]= \alpha \Hc(\pi(a_t|s_t))$. As aforementioned, the soft $Q$-function contains the sum of current and  future rewards and the sum of only future policy entropy. Since  we consider only the entropy terms without rewards here, the first term $\mathbb{E}_{a_t \sim \pi(\cdot|s_t)} [Q^{\pi_{old}}(s_t,a_t)]$ is the current estimate of the sum of future entropy when action $a_t$ is taken from policy $\pi$ at state $s_t$, whereas the second term $\alpha \Hc(\pi(a_t|s_t))$ is the entropy of the policy $\pi$ itself.   Hence, at each time step, SAC tries to update  the policy $\pi$ to yield the maximum sum of the estimated future entropy  and the policy entropy itself. Here, the term  $\mathbb{E}_{a_t \sim \pi(\cdot|s_t)} [Q^{\pi_{old}}(s_t,a_t)]$ plays the role of guiding  the policy towards the direction of high future entropy.

\begin{figure}[!t]
\centering
\begin{minipage}[b][0.585\textwidth][s]{0.27\textwidth}
\centering
\subfigure[Continuous 4-room maze]{\hspace{.6em}\includegraphics[height=0.5\textwidth,width=1.0\textwidth]{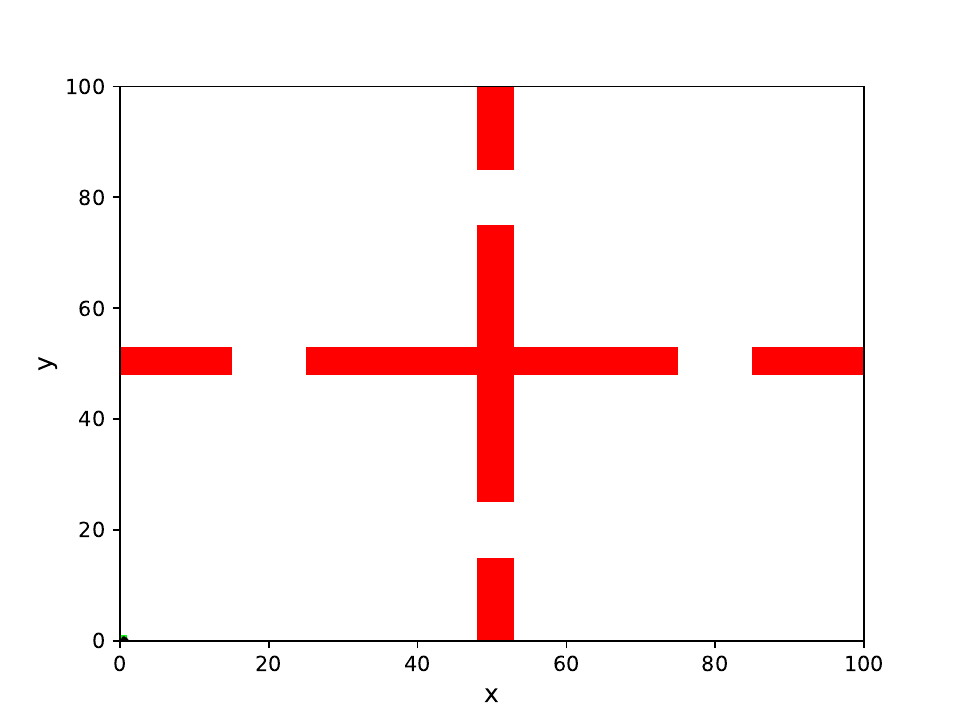}\label{fig:contmaze}}
\subfigure[Number of state visits]{\includegraphics[height=0.55\textwidth,width=\textwidth]{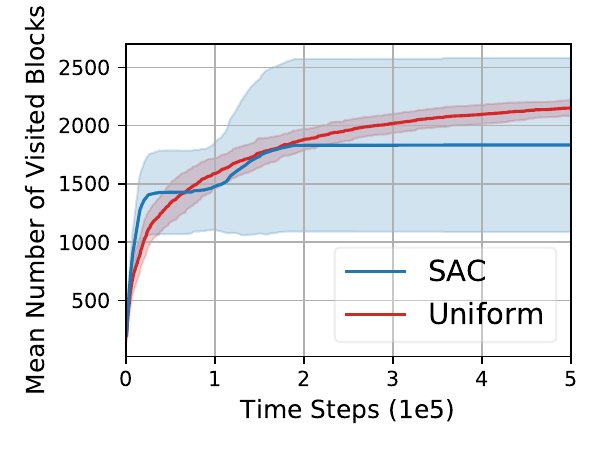}\label{fig:meanvisit}}
\subfigure[Gradient of $Q$-function]{\includegraphics[height=0.55\textwidth,width=\textwidth]{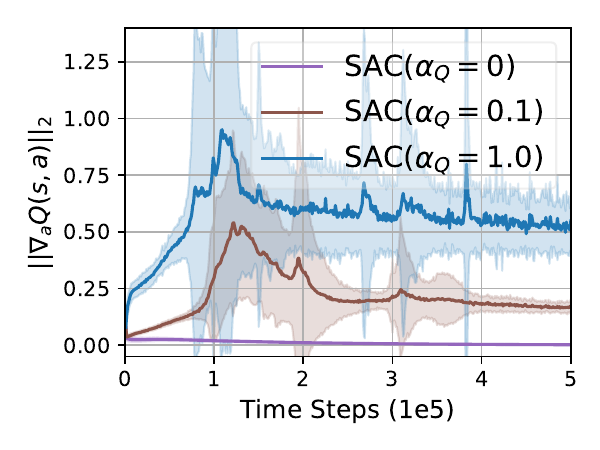}\label{fig:gradq}}
\end{minipage}
\hspace{1.5em}
\begin{minipage}[b][0.585\textwidth][s]{0.61\textwidth}
\centering
\subfigure[State histogram over 50k time steps starting from 300k, 350k, 400k, and 450k time steps (from the left  in order) ]{\includegraphics[height=0.86\textwidth,width=\textwidth]{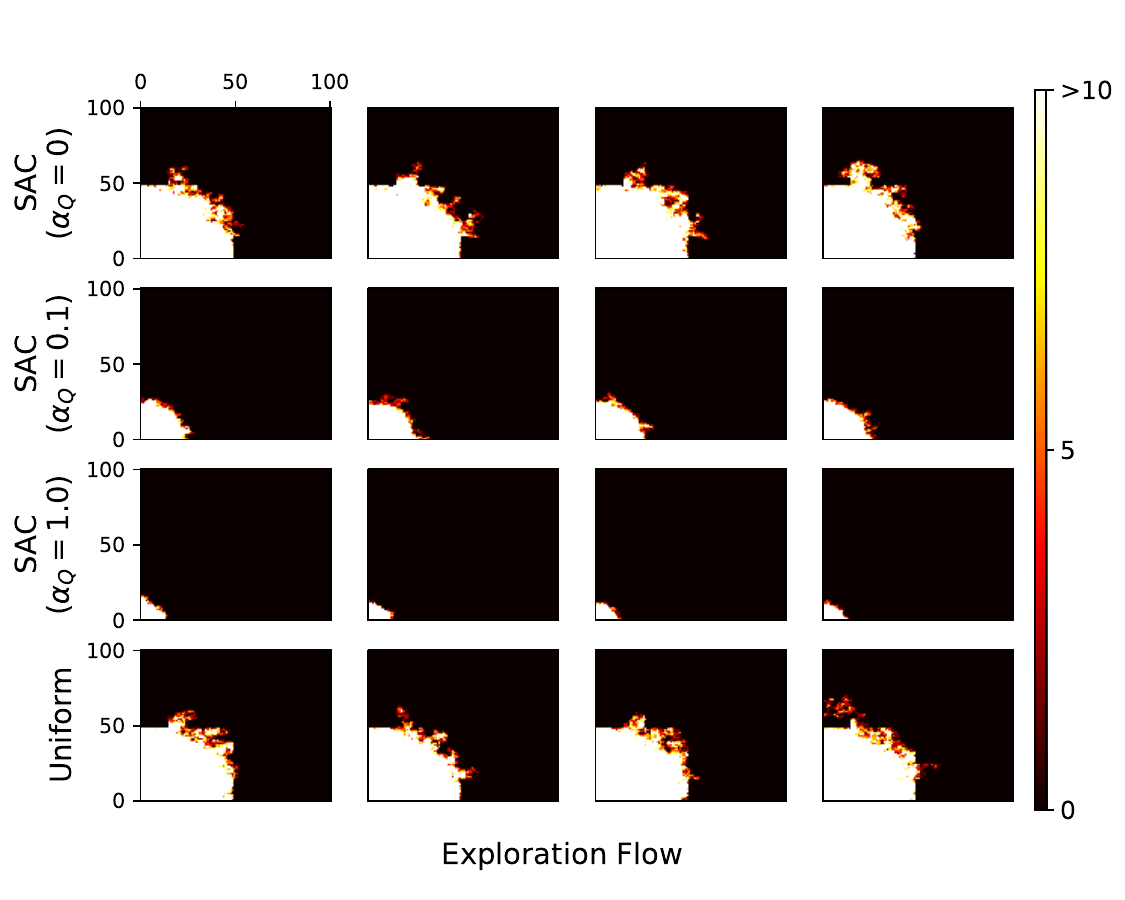}\label{fig:visitstate}}
\end{minipage}
\vspace{-3em}
\caption{Comparison of SAC and the uniform policy in the continuous 4-room maze}
\vspace{-1em}
\end{figure}

\textbf{Saturation:}  In sample-based update with function approximation, however, the SAC iteration does not yield the desired result, contrary to the intention behind maximum entropy. To see this, let us consider a pure exploration task in which there is no reward.   
 The considered task is a $100\times 100$ continuous 4-room maze proposed in \citep{han2020diversity}, modified from the continuous grid map available at \url{https://github.com/huyaoyu/GridMap}. Fig. \ref{fig:contmaze} shows the maze environment, where state is the $(x,y)$-position of the agent in the map,   action is $(dx,dy)$ bounded by $[-1,1]\times [-1,1]$, and  the next state of the agent is $(x+dx,y+dy)$. Starting from the left-lower corner $(0.5,0.5)$, the agent explores the maze without any external reward. 
First, note that for this pure exploration task, the optimal policy  maximizing $J_{MaxEnt}(\pi)$ is given by the uniform policy that selects all actions in  $\mathcal{A} = [-1,1]\times [-1,1]$ uniformly regardless of the value of $s_t$.  This is because the uniform distribution has maximum entropy for a bounded space \citep{cover1999elements}. Then, we compare the exploration behaviour of SAC and the uniform policy in the maze task. Fig. \ref{fig:meanvisit} shows the mean accumulated number of  different visited states averaged over $30$ random seeds as time goes, where the shaded region in the curve represents standard deviation (1$\sigma$) from the mean and a different state is meant as a nonoverlapping quantized $1\times 1$ square.  As seen in Fig. \ref{fig:meanvisit}, SAC explores more states than the uniform policy at the early stage of learning. As learning progresses, however, SAC fails to visit new states after $300k$ time steps, whereas the uniform policy continues visiting new states. As a result, SAC eventually visits fewer states than the uniform policy on average. The result shows that SAC fails to converge to the optimal uniform policy and its performance become saturated.

%\subsection{Narrow Exploration Radius due %to $Q$-function}
%\label{subsec:narrow}

\textbf{Narrow Exploration Radius:}   To examine the saturation behavior of SAC in the above pure exploration task, we investigate the policy update of SAC in \eqref{eq:sacpolup}.
Since the current $Q$-function estimate (implemented by a neural network)  replaces $Q^{\pi_{old}}$ in \eqref{eq:sacpolup} in implementation with function approximation, the policy update is rewritten  as 
\begin{equation}
\mathop{\arg\max}_{\pi\in\Pi} \{ \mathbb{E}_{a_t\sim\pi(\cdot|s_t)}[Q(s_t,a_t)] + \alpha\mathcal{H}(\pi(\cdot|s_t)) \}.
\label{eq:polupimp}
\end{equation}
As mentioned already, the first term $\mathbb{E}_{a_t \sim \pi(\cdot|s_t)} [Q(s_t,a_t)]$ is the current estimate of the sum of future entropy in this pure exploration case when action $a_t$ is taken from policy $\pi$ at state $s_t$, whereas the second term $\alpha \Hc(\pi(a_t|s_t))$ is the entropy of the policy $\pi$ itself.  The first term  $\mathbb{E}_{a_t \sim \pi(\cdot|s_t)} [Q(s_t,a_t)]$ intends to direct  the policy towards the direction of high future entropy. 
Note that maximizing the second term already yields the uniform policy, but the $Q$-function term affects the policy update.
In order to see how the $Q$-function term actually affects the policy update, we differentiate the entropy coefficient $\alpha$ in the policy update part \eqref{eq:sacpolup} or \eqref{eq:polupimp} as the policy entropy coefficient $\alpha_{\pi}$ and that in the soft value function part  \eqref{eq:softq} and \eqref{eq:bellman}
as the value entropy coefficient $\alpha_Q$. We fix $\alpha_\pi$ as $\alpha_{\pi}=1$ and change $\alpha_Q$ as  $0,~0.1,$ and $1$ (note that the case of $\alpha_Q=1$ is original SAC). With this change of $\alpha_Q$, we conducted the same pure exploration maze task. 
Fig. \ref{fig:gradq} shows the  average norm of the gradient of $Q$-function  with respect to action, i.e., $\mathbb{E}_{s_t \sim \mathcal{D}}[||\nabla_{a} Q(s_t,a)|_{a=a_t}||]$  over time with $a_t \sim \pi(\cdot|s_t)$ and $s_t$ from a mini-batch drawn from the replay buffer $\mathcal{D}$ of SAC update, where the $Q$ neural network weights were initialized randomly.  
Fig. \ref{fig:visitstate} shows the histogram of states that the policy visits over 50k time steps starting from 300k, 350k, 400k, and 450k time steps. When $\alpha_Q=0$ with no reward, the $Q$-function update by the Bellman operator $\mathcal{T}^\pi$ in  \eqref{eq:bellman} is trivial as $Q(s,a)\leftarrow \mathbb{E}_{s'\sim P(\cdot|s,a),a'\sim \pi(\cdot|s')} [Q(s',a')]$, i.e., replacement. When the initial $Q(s,a)$ is (nearly) flat over $\mathcal{S}\times \mathcal{A}$ by initial random weight assignment for the $Q$-neural network, the flat $Q$ is maintained by this trivial update. Indeed, it is seen in Fig. \ref{fig:gradq} that $\mathbb{E}_{s_t \sim \mathcal{D}}[||\nabla_{a} Q(s_t,a)|_{a=a_t}||]$ with $a_t \sim \pi(\cdot|s_t)$  is nearly zero across all time for $\alpha_Q=0$.   With 
a flat function $Q(s_t,\cdot)\approx c$ over the action space $\mathcal{A}$,  the first term 
 $\mathbb{E}_{a_t \sim \pi(\cdot|s_t)} [Q(s_t,a_t)]$ in \eqref{eq:polupimp} does not affect the policy update, only the second term $\Hc(\pi(\cdot|s_t))$ works, and thus the policy update yields $\pi$ to converge to the uniform policy for every state maximizing the total entropy. Hence,  the exploration radius in the case of $\alpha_Q=0$  is  almost the same as that of the uniform policy, as seen in Fig. \ref{fig:visitstate}.   When  $\alpha_Q>0$,  on the other hand,  the $Q$-function starts to be updated nontrivially by the Bellman operator  $\mathcal{T}^\pi$ in \eqref{eq:bellman} due to the $-\log \pi(a_{t+1}|s_{t+1})$ term in $V(s_{t+1})$ in \eqref{eq:bellman}, with $\pi$ given by the current policy.  
 It is now seen in Fig. \ref{fig:gradq} that $\mathbb{E}_{s_t \sim \mathcal{D}}[||\nabla_{a} Q(s_t,a)|_{a=a_t}||]$  is not zero anymore and the gradient norm becomes larger as $\alpha_Q$ increases from 0.1 to 1.0.
 Non-zero  $\mathbb{E}_{s_t \sim \mathcal{D}}[||\nabla_{a} Q(s_t,a)|_{a=a_t}||]$ means 
that  $Q(s_t,\cdot)$ as a function of action $a_t$ for given $s_t$ is not flat anymore and the first term in \eqref{eq:polupimp} affects the policy update so that the policy is updated for the direction of high $Q$-value (with intention for high future entropy) as well as high policy entropy $\mathcal{H}(\pi)$. As seen in Fig. \ref{fig:visitstate}, however,  the exploration radius reduces as $\alpha_Q$ increases from 0 to 1. The iteration process does not evolves for wider exploration as  intended.

%%%%%%%%%%%%%%%%%%%%%%%%%%%%%%%%%%%%%
\section{Methodology}
\label{sec:method}

\subsection{A Deeper Look at Pure Exploration} \label{sec:deep}

\begin{figure}
    \centering
    \subfigure[Number of state visits]{\includegraphics[width=0.27\textwidth]{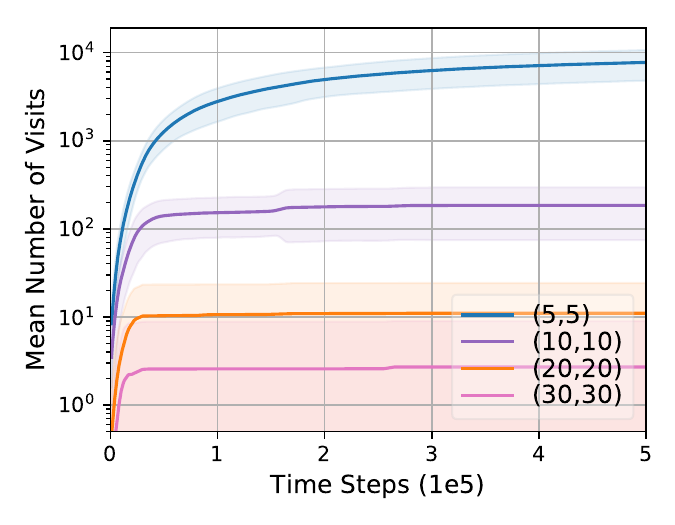}\label{fig:fairvisit}}
    \subfigure[$Q$-value difference]{\includegraphics[width=0.27\textwidth]{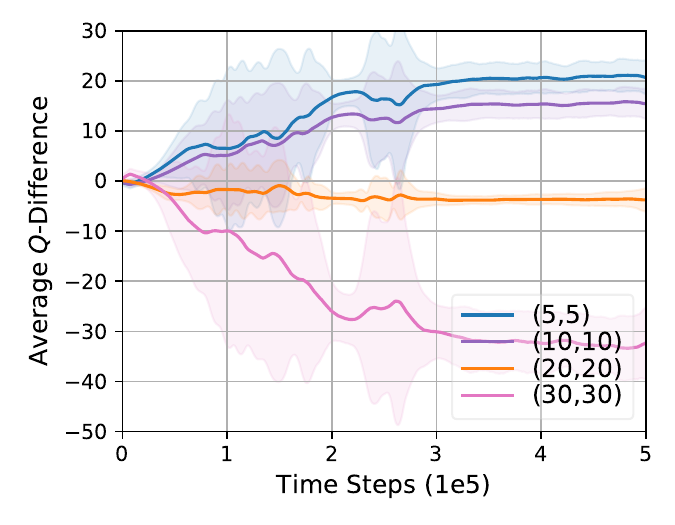}\label{fig:fairqf}}
    \subfigure[Empirical entropy]{\includegraphics[width=0.27\textwidth]{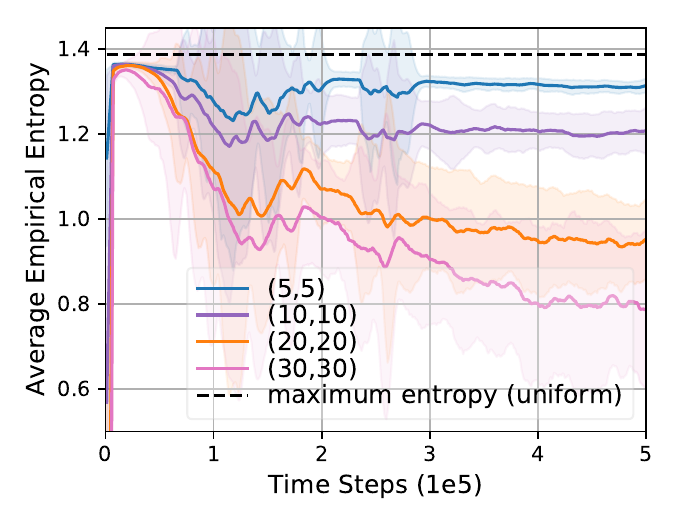}\label{fig:fairent}}
    \subfigure[Cross-section of $Q$-function and $\log\pi$ along the action line at square center]{\includegraphics[width=0.8\textwidth]{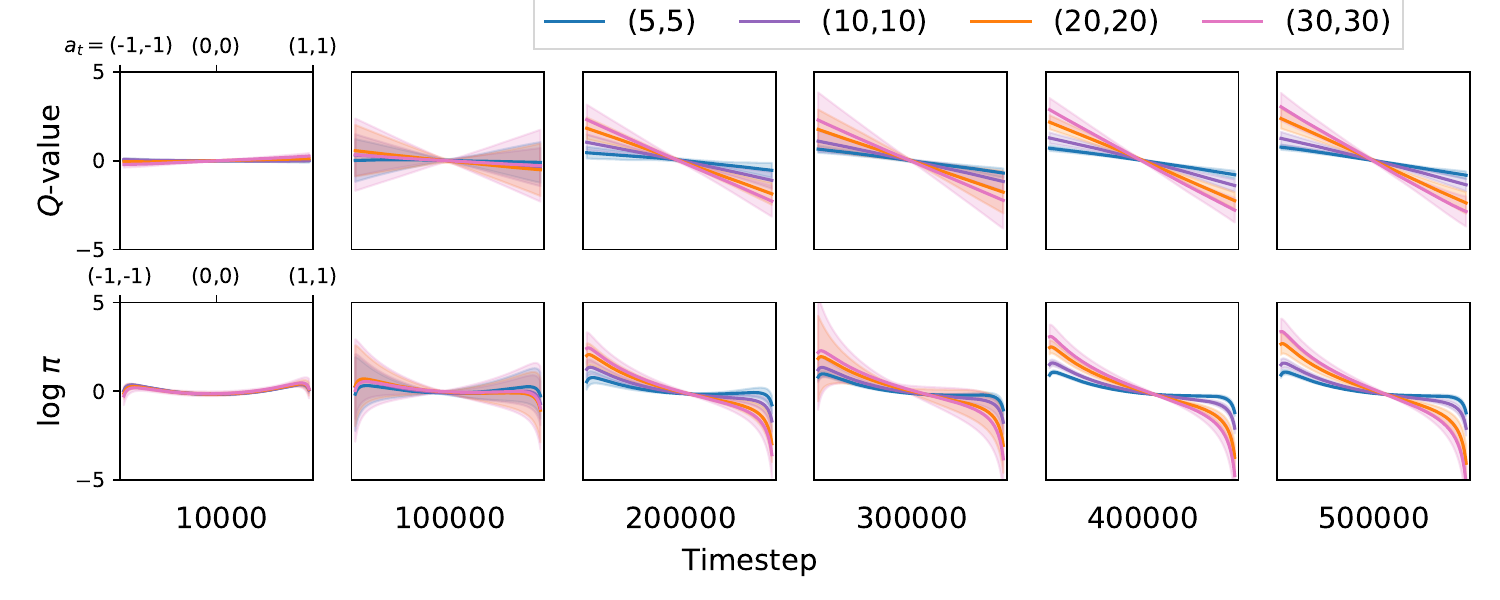}\label{fig:fairqp}}
    \vspace{-.5em}
    \caption{Analysis on sample trajectories of SAC in the continuous maze task}
    \label{fig:samptraj}
    \vspace{-1em}
\end{figure}

In order to propose our new approach overcoming the limitation of SAC implementation of  the maximum entropy framework, 
we first take a deeper look at  how the $Q$-function term  in \eqref{eq:polupimp}   hinders exploration, as SAC  (with $\alpha_{\pi}=\alpha_Q=1$) learns the maze task. For this, we consider four $2\times2$ squares centered at  $(5,5)$, $(10,10)$, $(20,20)$, and $(30,30)$ in the $100 \times 100$ maze, where every episode starts from (0.5,0.5). 
Fig. \ref{fig:fairvisit} shows the  number of accumulated  visits to  each square as time elapses. 
Figs. \ref{fig:fairqf} and  \ref{fig:fairent} show the  estimated $Q$ value and the average empirical entropy of each square, respectively, as time goes. For Fig. \ref{fig:fairqf}, every 1000 time steps, we sampled 1000 states uniformly from each square and an action from the current policy for each sampled state, and computed the Q-value average over the 1000 state-action samples for each square. Then, we computed the mean value of the four average values of the four squares. Fig. \ref{fig:fairqf} shows the average Q value of each square relative to the four-square mean value as time goes.  For Fig \ref{fig:fairent},  every 1000 time steps, we sampled 1000 states uniformly from each square and computed the average empirical entropy $\mathbb{E}_{s_t}[\mathbb{E}_{a_t\sim\pi(\cdot|s_t)}[-\log \pi(a_t|s_t)]]$ of the current policy $\pi$ at time $t$ averaged over the 1000 sampled states $\{s_t\}$ from each square.  
The upper row of Fig. \ref{fig:fairqp} shows the cross-section of the estimate $Q$-function $Q(s,a)$ along the diagonal action line from $(-1,-1)$ to $(1,1)$ at the center state $s$ of each square, as time goes, where each curve is shifted in $y$-axis so that the mean value averaged over samples along the action line is matched to zero in $y$-axis. 
The lower row of Fig. \ref{fig:fairqp} shows the value of $\log \pi(a|s)$ of the current policy $\pi$ at time step $t$  along the diagonal action line from $(-1,-1)$ to $(1,1)$ at the center state $s$ for each square as time goes, where the curve is shifted in $y$-axis to match the mean value to zero in $y$-axis.

First, note from Fig. \ref{fig:fairvisit} that the farther a state is from the starting point $(0.5,0.5)$, the less the agent visits the state, and the visitation difference is large. At the early stage of learning starting with random $Q$-network weight initialization and random policy-network weight initialization, there is little $Q$-value difference with respect to either state or action, as seen in Figs. \ref{fig:fairqf} and \ref{fig:fairqp}, so the entropy term is dominant in the policy update \eqref{eq:polupimp} and the policy entropy increases with the policy distribution approaching the uniform distribution, as seen in Fig. \ref{fig:fairent}.
As time goes, learning of the $Q$-function with  the Bellman backup   \eqref{eq:bellman} progresses. Basically, the Bellman backup \eqref{eq:bellman} with no reward adds $\Delta Q_t=\gamma\{\mathbb{E}_{a_{t+1}\sim \pi(\cdot|s_{t+1})}[Q(s_{t+1},a_{t+1})]+\alpha\mathcal{H}(\pi(\cdot|s_{t+1}))\}-Q(s_t,a_t)$ to $Q(s_t,a_t)$ for every $(s_t,a_t) \in \mathcal{S} \times \mathcal{A}$. However, this is approximated in practical RL.  In sample-based off-policy learning with function approximation, RL typically stores visited state-action pairs in the replay buffer $\mathcal{D}$ and the above Bellman backup is approximated as updating the $Q$-function by minimizing the loss $\mathbb{E}_{(s_t,a_t)\sim \mathcal{D}}[(Q(s_t,a_t)-Q^{target}(s_t,a_t))^2]$ based on a mini-batch uniformly drawn from the buffer.
Under this off-policy  learning with experience replay, when the initial $Q$-function is roughly flat and small, $\Delta Q_t$ soon becomes positive (the policy update increases the entropy of the visited states and $\Delta Q_t$ soon becomes positive  for the visited states), and hence the $Q$-values of frequently-visited states are updated more and thus have higher $Q$-values as seen in Fig.\ref{fig:fairqf}. This is because these states are stored more into $\mathcal{D}$ and sampled more  from $\mathcal{D}$ at mini-batch generation.
\begin{comment}
Furthermore, the increasing true soft $Q$-value of soft policy iteration also contributes to $Q$-value increase.
\end{comment}
Then, the initial $Q$-value difference biases the policy to visit the states with high $Q$-values more frequently than the states with low $Q$-values because the policy is updated to choose actions that maximizes the expectation of $Q$-value $\mathbb{E}_{a_t\sim\pi(\cdot|s_t)}[Q(s_t,a_t)]$ in the policy update \eqref{eq:polupimp}. This is evident in the Fig. \ref{fig:fairqp}, which shows the estimate of $Q$-function and the value of $\log\pi$ along the diagonal line. At the early stage of learning (10k time step in the figure), $Q$-function is roughly flat and the policy is almost close to the uniform distribution for the action line. As the time steps go on, the $Q$-values of actions close to $(-1,-1)$ becomes higher than the $Q$-value of actions near $(1,1)$ due to the off-policy learning with experience replay,  as explained above. Then, the policy is updated to choose actions with high $Q$-values more frequently to maximize the $Q$-value expectation, so the probability of choosing action $(-1,-1)$ towards the left-lower corner becomes higher than that of action $(1,1)$ for the opposite direction. As the policy distribution leans toward a certain action and becomes asymmetric away from uniformity, the policy entropy decreases further.  As seen in Fig. \ref{fig:fairent}, the speed of the policy entropy decrease varies depending on the $Q$-value difference along the action line in Fig. \ref{fig:fairqp}, and the policy entropy difference deepens the $Q$-value difference between states in Fig. \ref{fig:fairqf} because the $Q$-value estimates the policy entropy sum of future states. This positive feedback continues until saturation, as seen in Fig. \ref{fig:fairqf}, and it results in the narrow exploration radius in Fig. \ref{fig:visitstate} because the policy will be forced to visit states with high $Q$-values only. Note that this positive feedback reduces the policy entropy due to the $Q$-value difference, contrary to the intention behind maximum entropy.

\subsection{Max-Min Entropy RL}
\label{subsec:mmer}

In order to break the unwanted positive feedback loop occurring when implementing the maximum entropy framework (i.e., max-max entropy framework) in the previous subsection, we must reduce the policy entropy difference between states to reduce the $Q$-value difference between states in the feedback loop. For this, we aim to learn the $Q$-function so that the policy visits states with low entropy, and the policy update increases the policy entropy of these low-entropy states. Under this principle, we propose a new max-min entropy (MME) framework that aims to learn the $Q$-function to estimate the negative sum of policy entropy, while maintaining the policy entropy maximization term $\mathcal{H}(\pi(\cdot|s_t))$ in the policy update to increase the policy entropy of the visited states. 
Thus, we define the reversed soft $Q$-function $Q_R^\pi(s_t,a_t)$ for MME as 
\begin{equation}
    Q_R^{\pi}(s_t,a_t) : =  r_t +  \mathbb{E}_{\tau_{t+1}\sim\pi}\left[\sum_{l=t+1}^\infty\gamma^{l-t} (r_l - \alpha_Q \mathcal{H}(\pi(\cdot|s_l)))\right],
    \label{eq:rsoftq}
\end{equation}
whereas the original soft $Q$-function of SAC in \eqref{eq:softq} is given by
\[
Q^\pi(s_t,a_t) := r_t +  \mathbb{E}_{\tau_{t+1}\sim\pi}\left[\sum_{l=t+1}^\infty\gamma^{l-t} (r_l + \alpha_\pi \mathcal{H}(\pi(\cdot|s_l)))\right].
\]
Note that the original soft $Q$-function $Q^\pi$ adds the policy entropy to the reward and drives the policy to visit states with high entropy. On the other hand, our reversed soft $Q$-function subtracts the policy entropy from the reward and drives the policy to visit states with low entropy. In this sense, we call $Q_R$ as the ``reversed'' soft $Q$-function because it desires the reverse behavior of soft $Q$-function.

Then, $Q_R^\pi$ is estimated by a real-valued function $Q_R: \mathcal{S}\times \mathcal{A} \rightarrow \mathbb{R}$ based on a  Bellman operator $\mathcal{T}_R^\pi$:
\begin{comment}
%\footnote{We only consider pure exploration (i.e., zero reward function) in Section \ref{subsec:mmer}, but define $Q_R^\pi$ and $\mathcal{T}_r^\pi$ as containing the reward terms for future use.}
\end{comment}
\begin{equation}
    \mathcal{T}_R^\pi Q_R (s_t,a_t)= r_t + \gamma \mathbb{E}_{s_{t+1}\sim P(\cdot|s_t,a_t)}[V_R(s_{t+1})],
    \label{eq:rbellman}
\end{equation}
where $V_R(s_t) = \mathbb{E}_{a_t\sim \pi(\cdot|s_t)}[Q_R(s_t,a_t) + \alpha_Q \log \pi(a_t|s_t)]$ is the reversed soft state value function. At each iteration, the policy of MME is updated as
\begin{equation}
    \pi_{new} = \mathop{\arg\max}_{\pi\in\Pi}\mathbb{E}_{a_t\sim \pi(\cdot|s_t)}[Q_R^{\pi_{old}}(s_t,a_t)-\alpha_{\pi}\log\pi(a_t|s_t)],
\label{eq:mmerpolup}
\end{equation}
where $Q_R^{\pi_{old}}$ is substituted by the estimate function $Q_R$ at the iteration. 
Then, in pure exploration with no reward $r_t=0,\forall t$, the policy of MME will visit the states with low entropy due to the first term  $\mathbb{E}_{a_t\sim \pi(\cdot|s_t)}[Q_R^{\pi_{old}}(s_t,a_t)]$, and the policy entropy of those states will increase by the second term $\mathbb{E}_{a_t\sim \pi(\cdot|s_t)}[-\log\pi(a_t|s_t)] = \mathcal{H}(\pi(\cdot|s_t))$, as we intended. Note that the behaviour of the proposed method follows  the max-min  principle   \citep{chinchuluun2008pareto}, so we expect that our MME fairly increase the policy entropy of all states based on the fairness perspective of max-min optimization, whereas SAC increases  the policy entropy of states with high entropy only. The MME is expected to reduce the entropy difference and the $Q$-value difference between states to reduce the unwanted feedback loop and solve the saturation problem.
Furthermore, SAC considers the same entropy coefficient $\alpha_\pi$ for its policy update and the soft $Q$-function $Q^\pi$, but our MME distinguishes the policy entropy coefficient $\alpha_\pi$ in the policy update \eqref{eq:mmerpolup} and the value-entropy coefficient $\alpha_Q$ in the reversed soft $Q$-function $Q_R^\pi$ in \eqref{eq:rsoftq}, as we experimented in Section \ref{sec:motivation}. Changing $\alpha_Q$ and $\alpha_\pi$ allows for us to control the amount of the reversed $Q$-function in the policy update, and it will determine the ratio between the exploration due to the policy entropy and the exploration due to the reversed soft $Q$-function.

\begin{comment}
Comparing the MME policy update \eqref{eq:mmerpolup} with the policy update \eqref{eq:sacpolup} of SAC, we can view the objective function of the proposed MME RL as adding an intrinsic reward $r_t^{int} = -(\alpha_Q+\alpha) \gamma\mathbb{E}_{s_{t+1}\sim P(\cdot|s_t,a_t)}[\mathcal{H}(\pi_{old}(\cdot|s_{t+1}))]$ to the reward $r_t$ in the objective function of SAC  for better exploration, as in  intrinsic-motivated exploration methods \citep{achiam2017surprise, barto2013intrinsic, burda2018exploration}.
\end{comment}
In actual implementation, the negative entropy  in \eqref{eq:rsoftq} is plus-offsetted to make the $Q$-update increase the $Q$-value. 
The detailed implementation and algorithm of MME are provided in Appendix \ref{sec:detailedimp}.

\subsection{Disentangled Exploration and Exploitation for Rewarded Setup}
\label{subsec:disentangle}

In the previous subsection, we considered the problem from a pure exploration perspective. However, the ultimate goal of RL is to maximize the sum of rewards in rewarded environments, and the goal of exploration is to receive higher rewards without falling into local optima. With non-zero reward in \eqref{eq:rsoftq} - \eqref{eq:mmerpolup}, 
the policy will not only visit states with low entropy but also states with higher return.  In this case, the reward and the entropy are intertwined in the $Q$-function and then it is difficult to expect the intended MME exploration behavior through the intertwined $Q$-function.  Therefore, we disentangle exploration from  exploitation for rewarded setup, as considered in several previous works \citep{beyer2019mulex, simmons2019qxplore}, and propose disentangled MME (DE-MME) for rewarded setup. For this, we consider two policies: pure exploration policy $\pi_{E}$ that samples actions for pure exploration as described in Sec. \ref{subsec:mmer}, and target policy $\pi_{T}$ that actually interacts with the environment. 
We decompose the reversed soft $Q$-function $Q_R^\pi$ in \eqref{eq:rsoftq} into two terms $Q_R^\pi = Q_{R,R}^\pi + Q_{R,E}^\pi$, where $Q_{R,R}^\pi$ is the expected current and future reward sum considered in standard RL and $Q_{R,E}^{\pi}$ is the expected sum of future entropy:

\vspace{-1.4em}
{\footnotesize 
\[
   Q_{R,R}^\pi(s_t,a_t) = r_t + \mathbb{E}_{\tau_{t+1}\sim\pi}\left[\sum_{l=t+1}^\infty \gamma^{l-t} r_l\right],
    Q_{R,E}^\pi(s_t,a_t) =  \hspace{-0.2em}-\alpha_Q \mathbb{E}_{\tau_{t+1}\sim\pi}\left[\sum_{l=t+1}^\infty \gamma^{l-t} \mathcal{H}(\pi(\cdot|s_t))\right]. 
\]
}Then, we update the policy $\pi_{E}$ for pure exploration as
\begin{equation}
    \pi_{E,new}=\mathop{\arg\max}_{\pi'\in\Pi}\mathbb{E}_{a_t\sim\pi'(\cdot|s_t)}\left[Q_{R,E}^{\pi_{E,old}}(s_t,a_t)-\alpha_{\pi}\log\pi'(a_t|s_t)\right].
    \label{eq:poluppure}
\end{equation}
Note that increasing the expectation of $Q_{R,E}^{\pi_{E,old}}$ makes the policy visit states with low entropy of $\pi_{E}$, as we intended in the pure exploration case in Sec. \ref{subsec:mmer}. Finally,  we  update the target policy $\pi_{T}$ by using $Q_{R,E}^{\pi_{E,old}}$ as
\begin{equation}
    \pi_{T,new}=\mathop{\arg\max}_{\pi'\in\Pi}\mathbb{E}_{a_t\sim\pi'(\cdot|s_t)}\left[Q_{R,R}^{\pi_{T,old}}(s_t,a_t)+Q_{R,E}^{\pi_{E,old}}(s_t,a_t)-\alpha_{\pi}\log\pi'(a_t|s_t)\right].
    \label{eq:polupact}
\end{equation}
For implementation, $Q_{R,R}^{\pi_{T,old}}$ and $Q_{R,E}^{\pi_{E,old}}$ are  estimated by real-valued functions $Q_{R,R}$ and $Q_{R,E}$ based on their own Bellman operators (see Appendix \ref{sec:detailedimp}). 
Note  that the policy update \eqref{eq:mmerpolup} in Sec. \ref{subsec:mmer} can be expressed as maximizing   $\mathbb{E}_{a_t\sim\pi'(\cdot|s_t)}[Q_R^{\pi_{T,old}}(s_t,a_t) - \alpha_{\pi}\log \pi'(a_t|s_t)]$ over the target policy, where $Q_R^{\pi_{T,old}} =  Q_{R,R}^{\pi_{T,old}} + Q_{R,E}^{\pi_{T,old}}$. 
Thus, we can view that the policy update in \eqref{eq:polupact} replaces $Q_{R,E}^{\pi_{T,old}}$ in the previous policy update \eqref{eq:mmerpolup} with  $Q_{R,E}^{\pi_{E,old}}$  to disentangle exploration from exploitation.
In this way, the policy update \eqref{eq:polupact} will simultaneously increase the expectation of $Q_{R,R}^{\pi_{T,old}}$ to maximize the reward sum, the expectation of $Q_{R,E}^{\pi_{E,old}}$ to visit states with low entropy, and the policy entropy for diverse action.  The detailed implementation and algorithm for DE-MME are provided in Appendix \ref{sec:detailedimp}.

%%%%%%%%%%%%%%%%%%%%%%%%%%%%%%%%%%%%%%%%
\section{Experiments}
\label{sec:exp}
%%%%%%%%%%%%%%%%%%%%%%%%%%%%%%%%%%%%%%%%

We provide numerical results to show the performance of the proposed MME and DE-MME in pure exploration and various control
tasks. We provide source code for the proposed method at \url{http://github.com/seungyulhan/mme/} that requires Python Tensorflow. For all plots, the solid line represents the mean over random seeds and the shaded region represents 1 standard deviation  from the mean.

\begin{figure}
    \centering
    \subfigure[Number of visited states]{\includegraphics[height=0.36\textwidth,width=0.43\textwidth]{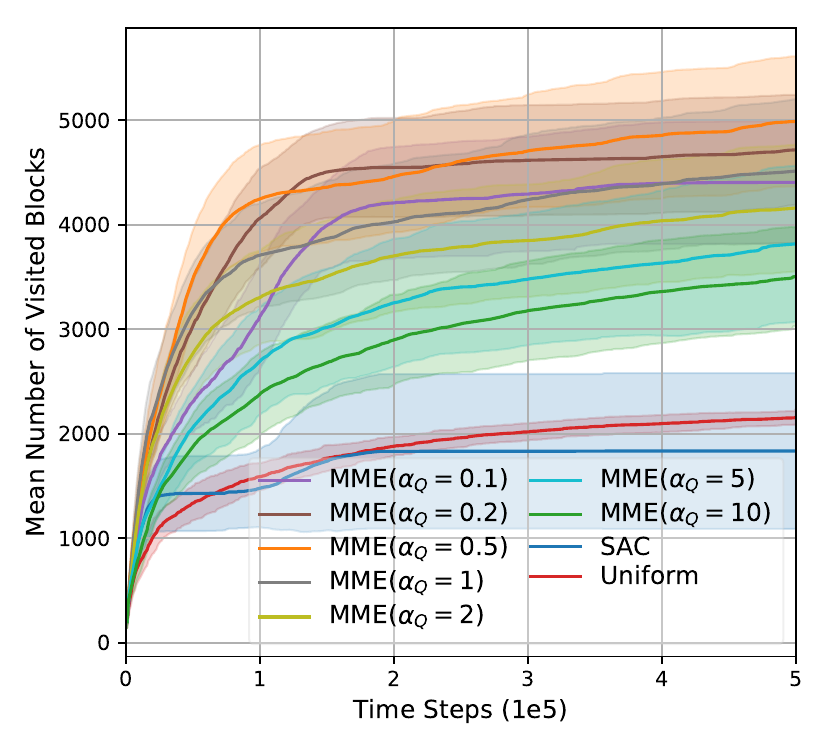}\label{fig:meanvisitmme}}
    \subfigure[State histogram of every 50k steps after 300k steps]{\includegraphics[height=0.395\textwidth,width=0.48\textwidth]{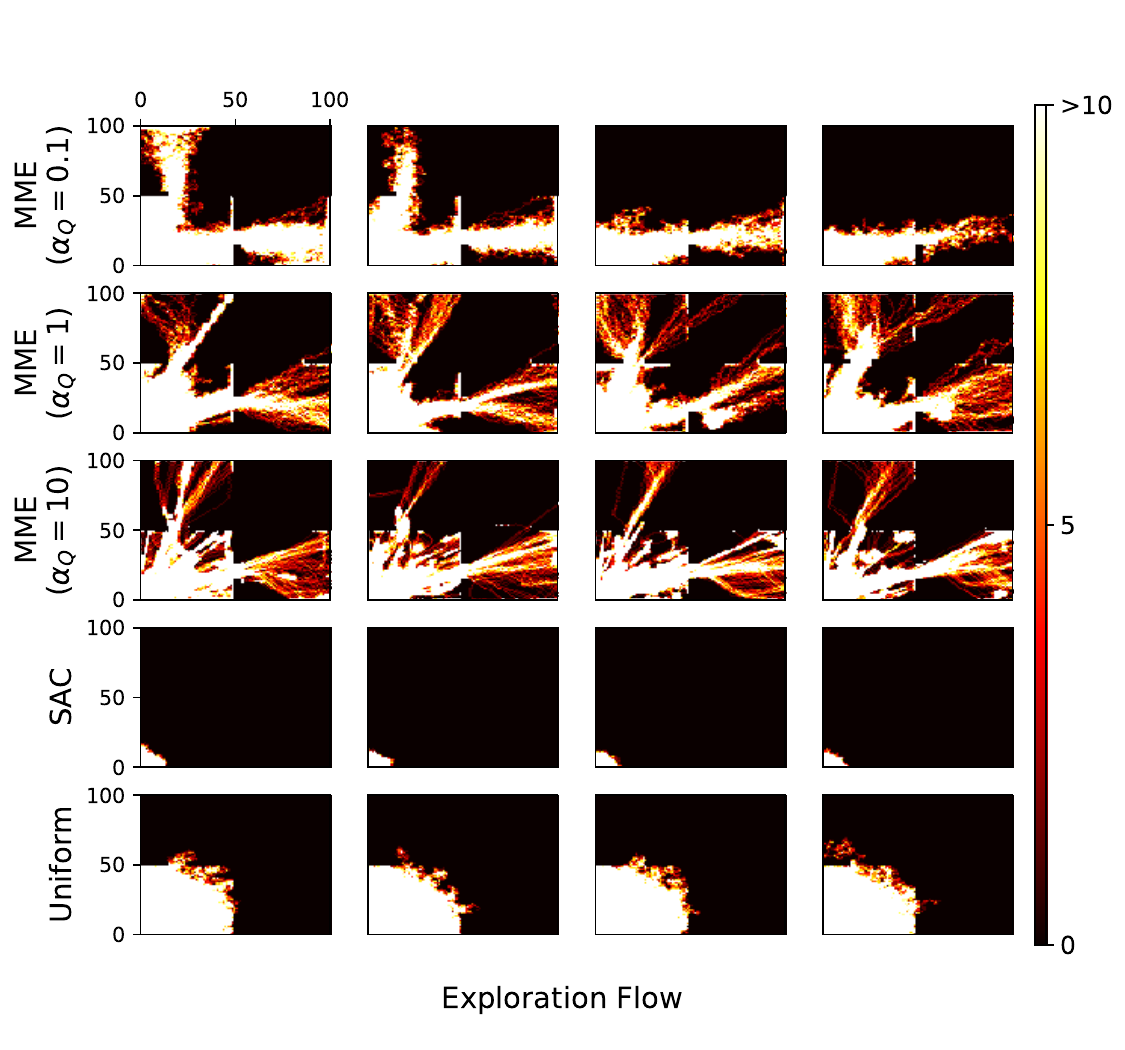}\label{fig:visitstatemme}}
    \vspace{-.5em}
    \caption{Comparison of MME (Proposed), SAC, and the uniform policy in the 4-room maze}
    \label{fig:contmazemme}
    \vspace{-1em}
\end{figure}

%%%%%%%%%%%%%%%%%%%%%%%%%%%%%%%%%%%%
\subsection{Pure Exploration}
\label{subsec:pureexp}

\begin{wrapfigure}{r}{0.6\textwidth}
\vspace{-1em}
    \centering
    \subfigure[$Q$-value difference]{\includegraphics[width=0.295\textwidth]{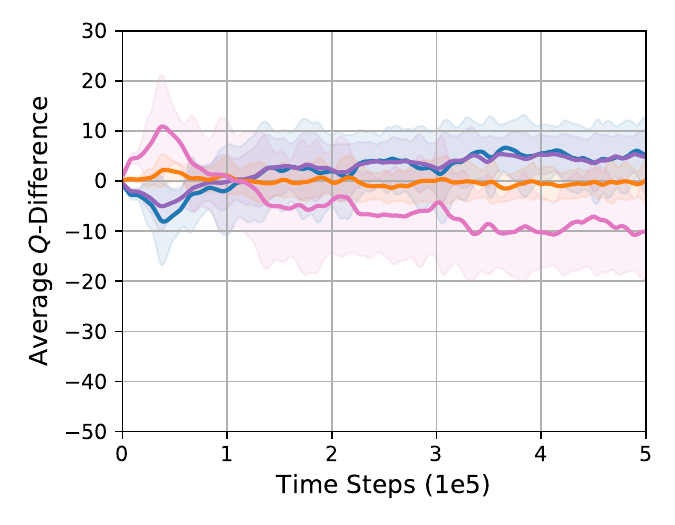}}
    \subfigure[Empirical entropy]{\includegraphics[width=0.295\textwidth]{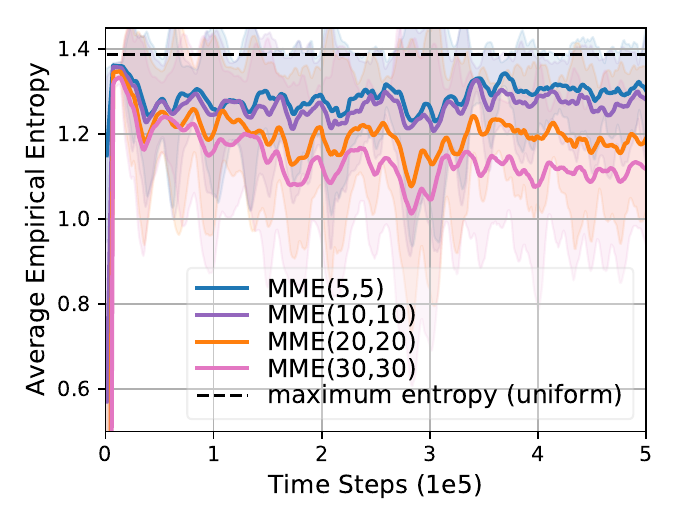}}
    \vspace{-.5em}
    \caption{Performance of MME}
    \label{fig:compentq}
\vspace{-1em}
\end{wrapfigure}
To see how the proposed method 
behaves in pure exploration, 
we considered the maze task described in Sec.\ref{sec:motivation} again. We compared  the exploration performance of MME in Sec.\ref{subsec:mmer}, SAC, and the uniform policy. For MME, we considered several $\alpha_Q\in\{0.1,0.2,0.5,1,2,5,10\}$ with $\alpha_{\pi}=1$.  Fig. \ref{fig:meanvisitmme} in the next page shows  the mean number of accumulated quantized visited states averaged over $30$ random seeds corresponding to Fig. \ref{fig:meanvisit}, and Fig. \ref{fig:visitstatemme} shows the histogram of visited states, of which setup is the same as Fig.\ref{fig:visitstate}. As seen in Fig. \ref{fig:meanvisitmme}, the proposed MME visits much more states than SAC or the uniform policy. In addition, we observe that MME continues discovering new states throughout the learning, while SAC rarely visits new states as learning progresses. As seen in Fig. \ref{fig:visitstatemme}, MME explores far and rare states as compared to SAC or the uniform policy, and this leads to a large enhancement  in exploration performance, as intended in Sec. \ref{subsec:mmer}. 
Note that the larger $\alpha_Q$ in update \eqref{eq:rsoftq}  \eqref{eq:mmerpolup} with $r_t=0$, the stronger is the effect of the $Q$-function term to visit states with low entropy and the weaker is the effect of the policy entropy term to explore widely in the action space, as we expected in Section \ref{subsec:mmer}. Hence, there is a trade-off between the two terms and $\alpha_Q=0.5$ seems best in the  maze task when $\alpha_\pi=1.0$, as seen in Fig. \ref{fig:meanvisitmme}. Thus, the result clearly shows why we  distinguish the policy entropy coefficient $\alpha_\pi$ and the value entropy coefficient $\alpha_Q$ for MME, whereas SAC uses the common entropy coefficient $\alpha=\alpha_\pi=\alpha_Q$.
We also plotted the $Q$-value difference and the empirical entropy of the four squares centered at (5,5), (10,10), (20,20) and (30,30) for MME, as done in Figs.
\ref{fig:fairqf} and
 \ref{fig:fairent}. The result is shown in  Fig. \ref{fig:compentq}. It is seen that the $Q$-value difference and the entropy difference among the states are clearly reduced  as compared to Figs.
\ref{fig:fairqf} and
 \ref{fig:fairent}.  It means that MME  breaks the unwanted positive feedback loop and improves the policy entropy of diverse states more uniformly as compared to SAC in terms of fairness under our max-min framework. This leads to better exploration, as seen in Fig \ref{fig:contmazemme}.

\begin{figure*}[!t]
	\centering
	\subfigure[SparseHopper-v1]{\includegraphics[width=0.24\textwidth]{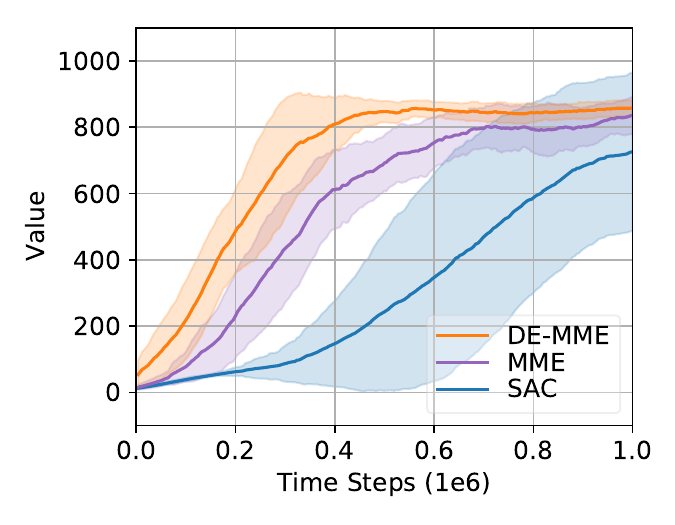}}
	\subfigure[SparseHalfCheetah-v1]{\includegraphics[width=0.24\textwidth]{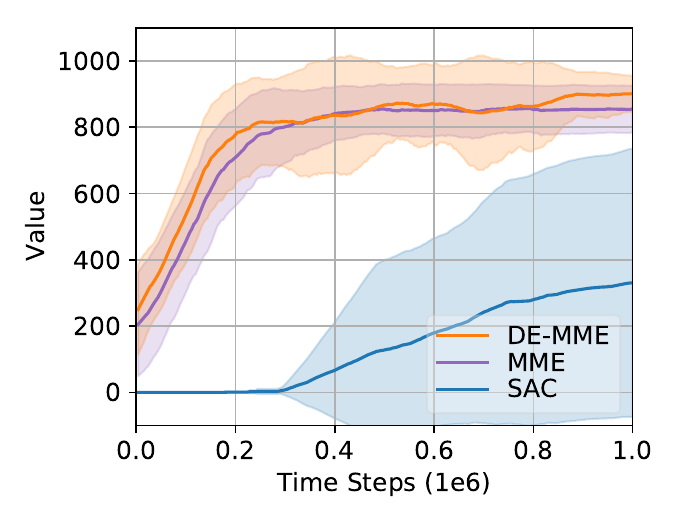}}
	\subfigure[SparseWalker2d-v1]{\includegraphics[width=0.24\textwidth]{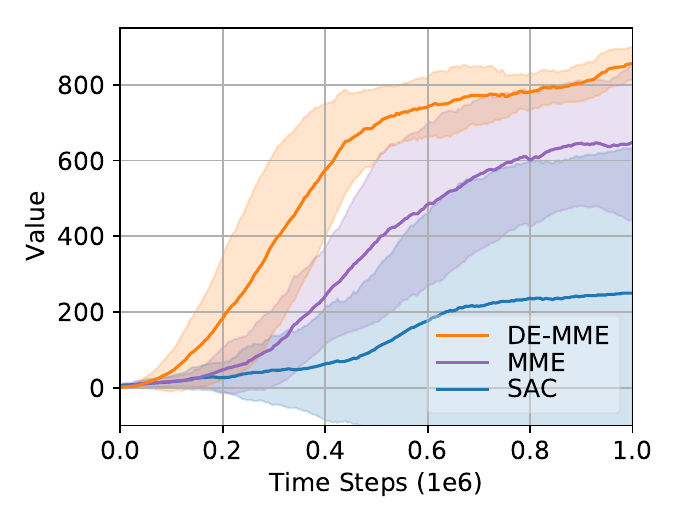}\label{fig:spswalker}}
	\subfigure[SparseAnt-v1]{\includegraphics[width=0.24\textwidth]{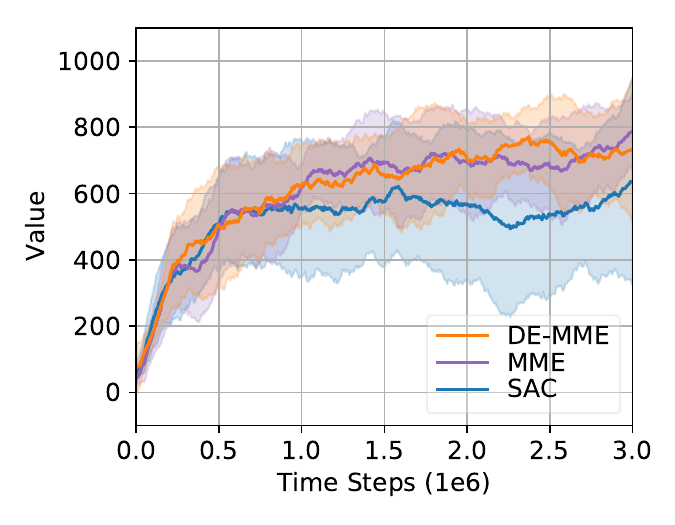}}
	\vspace{-0.5em}
	\caption{Performance comparison on Sparse Mujoco tasks}
	\label{fig:compsparsemujoco}
	\centering
	\subfigure[Del.Hopper-v1]{\includegraphics[width=0.24\textwidth]{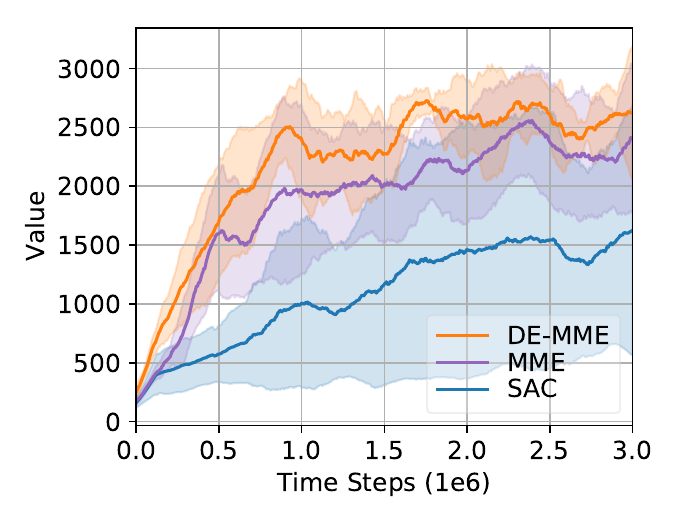}}
	\subfigure[Del.HalfCheetah-v1]{\includegraphics[width=0.24\textwidth]{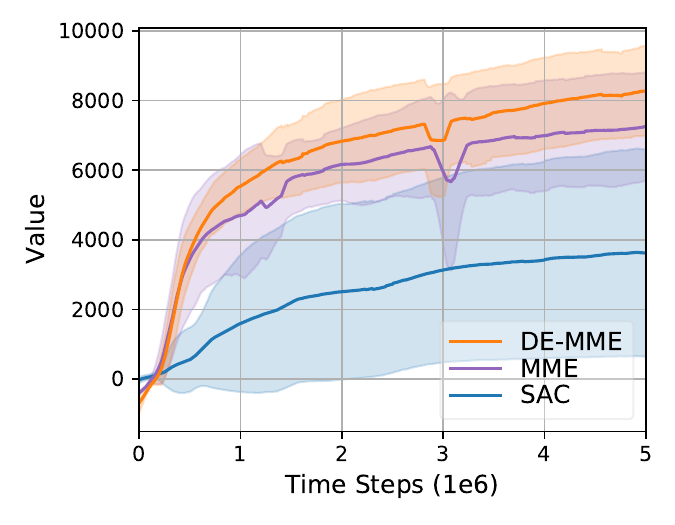}}
	\subfigure[Del.Walker2d-v1]{\includegraphics[width=0.24\textwidth]{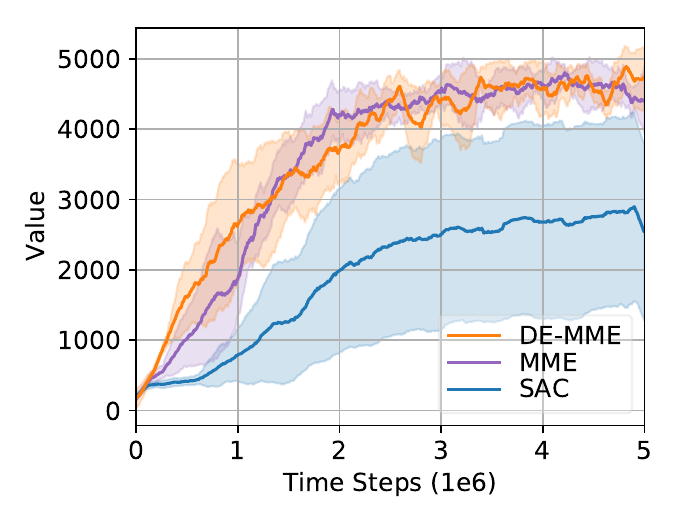}}
	\subfigure[Del.Ant-v1]{\includegraphics[width=0.24\textwidth]{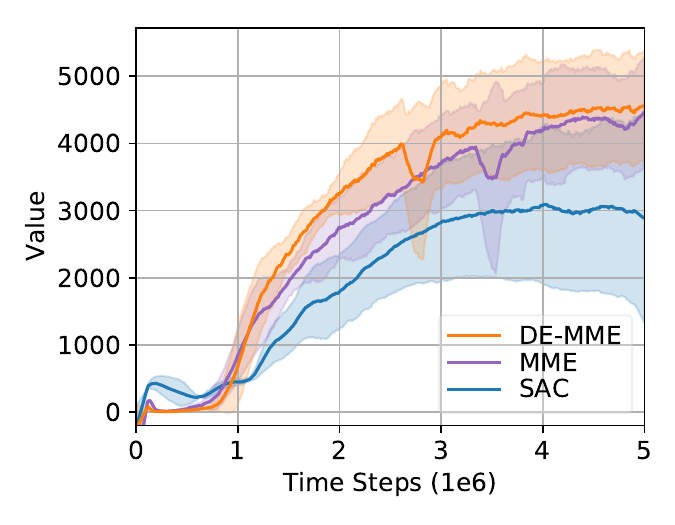}}
	\vspace{-0.5em}
	\caption{Performance comparison on Delayed Mujoco tasks}
	\label{fig:compdelayedmujoco}
    \vspace{-1.5em}
\end{figure*}

%%%%%%%%%%%%%%%%%%%%%%%%%%%%%%%%%%%%%%%
\subsection{Performance in Rewarded Environments}
\label{subsec:compsparse}

As mentioned in Sec.\ref{subsec:disentangle}, the ultimate goal of RL is to maximize the reward sum in rewarded environments and exploration is one of the means to achieve this goal.  Based on the enhanced exploration performance of MME, we expect  MME/DE-MME to show good performance in rewarded environments too.  
In order to verify this,  we considered three types of difficult control tasks for which current state-of-the-art RL algorithms do not show satisfactory performance:  Two types of sparse-reward tasks (SparseMujoco tasks and  DelayedMujoco tasks)  and high dimensional Humanoid tasks. 
SparseMujoco \citep{hong2018diversity,mazoure2019leveraging} is a sparse version of Mujoco \citep{todorov2012mujoco} in OpenAI Gym \citep{brockman2016openai}, and the reward is $1$ if the agent crosses the $x$-axis threshold $\tau$, otherwise $0$. DelayedMujoco \citep{guo2018generative, zheng2018learning} is a delayed version of Mujoco in which the reward is accumulated for $D$ time steps and the agent receives the accumulated reward sum once every $D$ time steps. During the accumulation time, the agent receives no reward. These sparse-reward environments have widely been  considered as challenging environments for validating the performance of exploration in many previous works  \citep{burda2018exploration, han2020diversity, hong2018diversity}.  

First, we compared the performance of MME algorithms to that of maximum entropy SAC in the sparse-reward tasks. For MME, we considered two versions: vanilla MME proposed in Sec.\ref{subsec:mmer}, and disentangled MME (DE-MME) proposed in Sec.\ref{subsec:disentangle}. For MME/DE-MME, we fixed  $\alpha_{\pi}$ of MME and DE-MME to be equal to $\alpha$ of SAC, and chose proper $\alpha_Q$ for each task. Detailed experimental setup is provided in Appendix \ref{sec:expsetup}. Figs. \ref{fig:compsparsemujoco} and  \ref{fig:compdelayedmujoco} show the  performance  averaged over $10$ random seeds on SparseMujoco tasks and $5$ random seeds on DelayedMujoco tasks, respectively. It is seen that  the proposed MME shows much higher performance than SAC in the considered environments with rewards. 
It is also seen that MME itself performs well enough in most environments but DE-MME indeed yields performance gain over vanilla MME and the gain is large in SparseWalker.
Thus,  disentanglement of exploration from exploitation is beneficial to MME for better reward performance in rewarded environments, as discussed in Sec.\ref{subsec:disentangle}. We provided the corresponding max average return tables in Appendix \ref{sec:addperfcomp} and ablation study for further analysis in Appendix \ref{sec:ablation}. There,  one of ablation study empirically shows that the performance enhancement by MME is caused by  improved exploration of MME as we intended.

\begin{comment}
To further analyze the performance results, we provide Fig. \ref{fig:meanvisitsw} that shows mean number of quantized state visits and Fig. \ref{fig:perfsw} that shows the resulting performance in SparseWalker2d, where the same $\alpha_Q=2$ is used for both MME and DE-MME. As seen in \ref{fig:meanvisitsw}, both DE-MME and MME visits much more states than SAC as in the pure exploration case, and it leads the performance enhancement compared to SAC in Fig. \ref{fig:perfsw}, as we expected. Surprisingly, MME visits more states than DE-MME, but MME has lower performance than DE-MME. 
\begin{wrapfigure}{r}{0.5\textwidth}
    \vspace{-1em}
    \centering
    \subfigure[Number of state visits]{\includegraphics[width=0.24\textwidth]{figures/_comp_envs/SparseWalker2d-v1_cov.pdf}\label{fig:meanvisitsw}}
    \subfigure[Performance]{\includegraphics[width=0.24\textwidth]{figures/_comp_envs/SparseWalker2d-v1_sw.pdf}\label{fig:perfsw}}
    \vspace{-.5em}
    \caption{Analysis on the SparseWalker2d task}
    \vspace{-1em}
\end{wrapfigure}
\end{comment}

Finally, we  compared the performance of MME/DE-MME to that of 
popular general RL algorithms and recent exploration methods 
on the considered sparse-reward environments (SparseMujoco and DelayedMujoco tasks) and 
dense-reward high-dimensional Mujoco tasks (Humanoid, HumanoidStandup). 
We considered several action-based exploration methods: SAC combined with divergence \citep{hong2018diversity} (SAC-Div) and diversity actor-critic (DAC) \citep{han2020diversity}, and  state-based exploration methods with random network distillation (RND) \citep{burda2018exploration} and MaxEnt (State) \citep{hazan2019provably}. For general RL algorithms, we considered several on-policy RL algorithms: proximal policy optimization (PPO) \citep{schulman2017proximal} and trust-region policy optimization (TRPO) \citep{schulman2015trust}, and entropy-based off-policy RL algorithms: soft Q-learning (SQL) \citep{haarnoja2017rein} and SAC \citep{haarnoja2018soft}. We provided detailed explanation and implementation for each algorithm in Appendix \ref{sec:addperfcomp}. 
Table \ref{table:marother} summarizes  the max average return result. It is seen that  MME/DE-MME have superior performance to other methods.

\begin{table*}[!h]
	\centering
	\begin{adjustbox}{width=1\textwidth}
		\begin{tabular}{L{7.1em}|C{7.5em}C{7.5em}C{7.5em}C{7.5em}C{7.5em}C{7.5em}}
			\hline
			& MME & DE-MME & DAC & SAC-Div & RND & MaxEnt(State)\\
			\hline
			Sps.Hopper & {\bf902.50$\pm$4.36} & 893.30$\pm$6.72 & {\bf900.30$\pm$3.93} & 817.40$\pm$253.54 & {\bf897.90$\pm$6.06} & 879.50$\pm$30.96 \\
			Sps.HalfCheetah & 903.50$\pm$34.97 & {\bf924.90$\pm$39.57} & {\bf915.90$\pm$50.71} & 394.70$\pm$405.53 & 827.80$\pm$85.61 & {\bf924.70$\pm$24.44}\\
			Sps.Walker2d & 818.00$\pm$208.60 & {\bf886.60$\pm$25.77} & 665.10$\pm$355.66 & 278.50$\pm$398.23 & 750.90$\pm$179.09 & 705.30$\pm$274.88 \\
			Sps.Ant  & 953.70$\pm$28.39 & {\bf973.60$\pm$12.55} & 935.80$\pm$37.08 & 870.70$\pm$121.14 & 920.60$\pm$107.50 & 940.70$\pm$43.84 \\
			Del. Hopper & {\bf3421.32$\pm$88.29} & {\bf3435.28$\pm$39.55} & {\bf3428.18$\pm$69.08} & 2090.64$\pm$1383.83 & 2721.06$\pm$1199.20 & 3254.10$\pm$30.75 \\
			Del. HalfCheetah & 7299.28$\pm$1562.19 & {\bf8451.20$\pm$1375.27} & 7594.70$\pm$1259.23 & 4080.67$\pm$3418.07 & 7429.94$\pm$1383.75 & 7907.98$\pm$535.41 \\
			Del. Walker2d & 5148.58$\pm$193.78 & {\bf5274.89$\pm$186.35} & 4067.11$\pm$257.81 & 4048.11$\pm$290.48 & 4098.63$\pm$683.36 & 4430.61$\pm$347.02 \\
			Del. Ant & 4664.04$\pm$836.37 & {\bf4851.64$\pm$830.88} & 4243.19$\pm$795.49 & 3978.34$\pm$1370.23 & 1361.36$\pm$704.69 & 1156.61$\pm$112.40 \\
			\hline
			& MME & DE-MME & SAC & SQL & PPO & TRPO \\
			\hline
			HumanoidStandup & {\bf267734.03 $\pm$74302.99} & 250935.53 $\pm$49386.43 & 167394.36 $\pm$7291.99 & 138996.84 $\pm$33903.03 & 160211.90 $\pm$3268.37 & 153919.84 $\pm$1575.62 \\
			Humanoid & {\bf9080.54$\pm$768.52} & 8607.75$\pm$570.61 & 6760.81$\pm$267.78 & 5010.72$\pm$248.59 & 6153.54$\pm$246.95 & 5730.74$\pm$455.90 \\
			\hline
		\end{tabular}
	\end{adjustbox}
	\caption{Max average return of MME/DE-MME and other recent RL algorithms}
	\label{table:marother}
	\vspace{-1em}
	%\vspace{3em}
\end{table*}
%add mar table?

% pure exploration / sparse -> SAC, SAC-Div, DAC, RND, MaxEnt
% delay / original -> SAC, etc

%%%%%%%%%%%%%%%%%%%%%%%%%%%%%%%%%%%%%%%%%%%
\section{Conclusion}
%%%%%%%%%%%%%%%%%%%%%%%%%%%%%%%%%%%%%%%%%%%
\label{sec:conclusion}

In this paper, we have proposed a MME framework for RL to resolve the unwanted exploration behavior of maximum entropy RL in off-policy learning with function approximation. In pure exploration, to implement MME, we train the $Q$-function to visit states with low entropy contrary to the maximum entropy strategy, while maintaining the policy entropy maximization term in the policy update. Then, we extended MME to rewarded environments. In rewarded environments we disentangle exploration from exploitation for MME to explore diverse states as in pure exploration as well as to achieve high return. Numerical results show that the proposed MME explores farther and wider in the state space than maximum entropy realization,  alleviates possible positive feedback of off-policy maximum entropy learning, and yields a significant enhancement in exploration and final performance over existing RL methods in various difficult tasks. As for potential impacts, RL can be applied to sensitive areas that require control, such as drone control. However, it is only a risk that RL itself has, and it is not very relevant to the work that we are trying to address in this paper.

%%%%%%%%%%%%%%%%%%%%%%%%%%%%%%%%%%%%%%%%%%%
\section{Acknowledgement}
%%%%%%%%%%%%%%%%%%%%%%%%%%%%%%%%%%%%%%%%%%%
\label{sec:ack}
This work is supported by Center for Applied Research in Artificial
Intelligence (CARAI) grant funded by Defense Acquisition Program  
Administration (DAPA) and Agency for Defense Development (ADD) (UD190031RD). Dr. Seungyul Han is currently with AI Graduate School of UNIST and his work is partly supported by Artificial Intelligence Graduate School support (UNIST),  Institute of Information \& Communications Technology Planning \& Evaluation (IITP) grant funded by the Korea government (MSIT) (No.2020-0-01336).

\newpage

\bibliography{referenceBibs}
\bibliographystyle{plain}

%%%%%%%%%%%%%%%%%%%%%%%%%%%%%%%%%%%%
\newpage
\appendix
\counterwithin{table}{section}
\counterwithin{figure}{section} 
\renewcommand{\theequation}{\thesection.\arabic{equation}}

\section{Detailed Implementation and Algorithm for Max-Min Ent RL}
\label{sec:detailedimp}

Here, we provide the detailed implementation and algorithm of MME proposed in Section \ref{sec:method}. 

\subsection{Detailed Implementation of MME}

First, we consider the vanilla MME proposed in Section \ref{subsec:mmer}. We approximate the policy $\pi$,  the reversed action value function $Q_R$, and the reversed state value function $V_R$ by using deep neural networks with parameters $\theta$, $\phi$, and $\psi$, respectively. Based on the policy update of MME in \eqref{eq:mmerpolup}, we define the practical objective function $\hat{J}_\pi(\theta)$ for the parameterized policy $\pi_\theta$, given by
\begin{equation}
    \hat{J}_\pi(\theta) = \mathbb{E}_{s_t\sim\mathcal{D},a_t\sim\pi_\theta(\cdot|s_t)}[Q_{R,\phi}(s_t,a_t)-\alpha_{\pi}\log \pi_\theta(a_t|s_t)].
    \label{eq:pracpol}
\end{equation}
Furthermore, based on the Bellman operator $\mathcal{T}_R^\pi$ in \eqref{eq:rbellman}, we define the practical loss functions $\hat{L}_{Q_R}(\phi)$ and $\hat{L}_{V_R}(\psi)$ for the parameterized reversed value functions $Q_{R,\phi}$ and $V_{R,\psi}$, respectively, given by
\begin{align}
    \hat{L}_{Q_R}(\phi)&=\mathbb{E}_{(s_t,a_t)\sim\mathcal{D}}\left[\frac{1}{2} (Q_{R,\phi}(s_t,a_t) - \hat{Q}(s_t,a_t))^2\right],\\
    \hat{L}_{V_R}(\psi)&=\mathbb{E}_{(s_t,a_t)\sim\mathcal{D}}\left[\frac{1}{2} (V_{R,\psi}(s_t) - \hat{V}(s_t))^2\right],
\end{align}
where the target values $\hat{Q}$ and $\hat{V}$ are defined as
\begin{align}
    \hat{Q}(s_t,a_t) &= r_t + \gamma \mathbb{E}_{s_{t+1}\sim P(\cdot|s_t,a_t)}[V_{R,\bar{\psi}}(s_{t+1})],\label{eq:pracrq}\\
    \hat{V}(s_t) &= \mathbb{E}_{a_t\sim \pi_\theta(\cdot|s_t)}[ Q_{R,\phi}(s_t,a_t) + \alpha_Q\log\pi_\theta(a_t|s_t)].\label{eq:pracrv}
\end{align}
For implementation, we divide $r_t$ by $\alpha_{\pi}$ (scaling reward by $1/\alpha_{\pi}$) in \eqref{eq:pracrq} instead of multiplying $\alpha_{\pi}$ by $\log\pi_\theta$ in \eqref{eq:pracpol}, similarly to the implmentation of SAC \citep{haarnoja2018soft}.  $\bar{\psi}$ denotes the neural network parameter of the target value $V_{\bar{\psi}}$ for stable learning, and it is updated by exponential moving average (EMA) of $\psi$ \citep{mnih2015human}.

\begin{figure}[h]
	\centering
	\vspace{-4em}
	\includegraphics[width=0.9\textwidth]{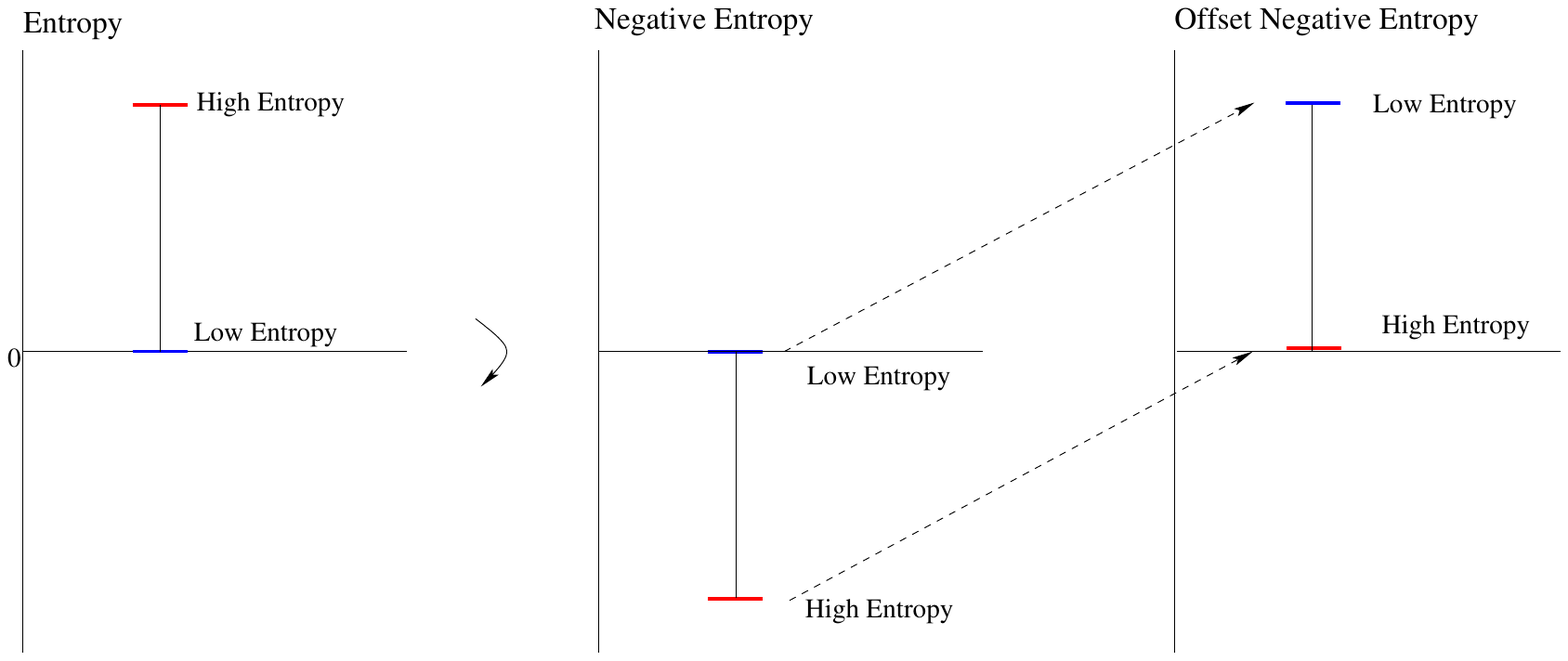}
	\vspace{-5.5em}
	\caption{Offset Negative Entropy}
	\label{fig:entropyoffset}
\end{figure}

Note that in the reversed value function  \eqref{eq:pracrv}, the negative entropy $\mathbb{E}_{a_t \sim \pi_\theta(\cdot|s_t)}\log \pi_\theta(a_t|s_t)$ is added so that lower entropy states yield higher $Q$-values. Note that the entropy is not necessarily positive in continuous space, but the policy entropy becomes positive soon due to the entropy maximization term $\mathcal{H}(\pi)$ in the policy update, so the negative entropy becomes negative. Hence, if this negative entropy is directly added to $Q$, the $Q$-value can decrease in the early stage of learning since $Q$-function starts near $0$ from the random initialization of $Q$-function network. Note that MME breaks the feedback loop assuming that $Q$-function continues to increase, so the decrease of $Q$-function may result in unpredictable effect on learning.
\begin{comment}So, we prefer actual larger increase in $Q$ by lower entropy to smaller decrease in $Q$ by lower entropy.\end{comment}
Hence,  we add a positive offset to the negative entropy in \eqref{eq:pracrv}  so that the offset negative entropy is positive, as seen in Fig.	\ref{fig:entropyoffset}.
We considered two types of positive offset: 1) maximum entropy constant $D_a \log 2$ (because the action space is bounded as $[-1,1]$) and 2) $-\min_m\log \pi_\theta(a_m|s_m)$ in mini-batch $\{(s_m,a_m), ~m=1,\cdots,M\}$ sampled from the buffer. In both cases, the offset negative entropy is positive, but the latter case showed better exploration. Hence,  we considered the latter case throughout this paper.

In addition, we considered two $Q$-functions $Q_{R,\phi_i},~i=1,2$ and the minimum of two $Q$-functions was used for the policy update in \eqref{eq:pracpol} to reduce overestimation bias as proposed in \citep{fujimoto2018addressing}. Each $Q_{R,\phi_i}$ minimizes its own loss function $\hat{L}(\phi_i)$ with the common $V_{R,\psi}$. This technique was also considered in SAC.

\subsection{Detailed Implementation of DE-MME}

The implementation of  the disentangled version of MME (DE-MME) proposed in Section \ref{subsec:disentangle} is as follows. In DE-MME, we have two policies $\pi_T$ and $\pi_E$ parameterized  with parameters $\theta_{T}$ and $\theta_{E}$, respectively.  $Q_R^\pi$ is decomposed into $Q_{R,R}^\pi$ and $Q_{R,E}^\pi$, and   $Q_{R,R}^{\pi_{T,old}}$ and $Q_{R,E}^{\pi_{E,old}}$ are estimated by  two estimate $Q$-functions $Q_{R,R}$ and $Q_{R,E}$ by using their respective Bellman operators $\mathcal{T}_{R,R}^{\pi_{T,old}}$ and $\mathcal{T}_{R,E}^{\pi_{E,old}}$,  given by
\begin{align}
\mathcal{T}_{R,R}^{\pi_{T,old}} Q_{R,R}(s_t,a_t) &= r_t + \mathbb{E}_{s_{t+1}\sim P(\cdot|s_t,a_t)}[V_{R,R}(s_{t+1})]\label{eq:rewbellman}\\
\mathcal{T}_{R,E}^{\pi_{E,old}} Q_{R,E}(s_t,a_t) &=  \mathbb{E}_{s_{t+1}\sim P(\cdot|s_t,a_t)}[V_{R,E}(s_{t+1})],\label{eq:purebellman}
\end{align}
where $V_{R,R}(s_t) = \mathbb{E}_{a_t\sim \pi_{T,old}(\cdot|s_t)}[Q_{R,R}(s_t,a_t)]$ and $V_{R,E}(s_t) = \mathbb{E}_{a_t\sim \pi_{E,old}(\cdot|s_t)}[Q_{R,E}(s_t,a_t)+\alpha_Q\log\pi_{E,old}(a_t|s_t)]$. For implementation, we parameterize the value functions $Q_{R,R}$, $V_{R,R}$, $Q_{R,E}$, $V_{R,E}$ by neural network parameters $\phi_R$, $\psi_R$, $\phi_E$, $\psi_E$, respectively. Then, based on the policy updates \eqref{eq:polupact} and \eqref{eq:poluppure}, we define the objective functions $\hat{J}_{\pi_T}(\theta_T)$ and $\hat{J}_{\pi_E}(\theta_E)$ respectively for the parameterized policies $\pi_{T,\theta_T}$ and $\pi_{E,\theta_E}$, given by
\begin{align}
    \hat{J}_{\pi_T}(\theta_T) &= \mathbb{E}_{s_t\sim\mathcal{D},a_t\sim\pi_{T,\theta_T}(\cdot|s_t)}[Q_{R,R,\phi_R}(s_t,a_t)+Q_{R,E,\phi_E}(s_t,a_t)-\alpha_{\pi}\log \pi_{T,\theta_T}(a_t|s_t)],\label{eq:dispracpolact}\\
    \hat{J}_{\pi_E}(\theta_E) &= \mathbb{E}_{s_t\sim\mathcal{D},a_t\sim\pi_{E,\theta_E}(\cdot|s_t)}[Q_{R,E,\phi_E}(s_t,a_t)-\alpha_{\pi}\log \pi_{E,\theta_E}(a_t|s_t)].\label{eq:dispracpolpure}
\end{align}
Based on the Bellman operators \eqref{eq:rewbellman} and \eqref{eq:purebellman}, we define the loss functions $\hat{L}_{Q_{R,R}}(\phi_R)$, $\hat{L}_{V_{R,R}}(\psi_R)$, $\hat{L}_{Q_{R,E}}(\phi_E)$, and $\hat{L}_{V_{R,E}}(\psi_E)$ for the parameterized value functions $Q_{R,R,\phi_R}$, $V_{R,R,\psi_R}$, $Q_{R,E,\phi_E}$, and $V_{R,E,\psi_E}$, respectively, given by
\begin{align}
    \hat{L}_{Q_{R,R}}(\phi_R)&=\mathbb{E}_{(s_t,a_t)\sim\mathcal{D}}\left[\frac{1}{2} (Q_{R,R,\phi_R}(s_t,a_t) - \hat{Q}_{R,R}(s_t,a_t))^2\right],\\
    \hat{L}_{V_{R,R}}(\psi_R)&=\mathbb{E}_{(s_t,a_t)\sim\mathcal{D}}\left[\frac{1}{2} (V_{R,R,\psi_R}(s_t) - \hat{V}_{R,R}(s_t))^2\right],\\
    \hat{L}_{Q_{R,E}}(\phi_E)&=\mathbb{E}_{(s_t,a_t)\sim\mathcal{D}}\left[\frac{1}{2} (Q_{R,E,\phi_E}(s_t,a_t) - \hat{Q}_{R,E}(s_t,a_t))^2\right],\\
    \hat{L}_{V_{R,E}}(\psi_E)&=\mathbb{E}_{(s_t,a_t)\sim\mathcal{D}}\left[\frac{1}{2} (V_{R,E,\psi_E}(s_t) - \hat{V}_{R,E}(s_t))^2\right],
\end{align}
where the target values $\hat{Q}_{R,R}$, $\hat{V}_{R,R}$, $\hat{Q}_{R,E}$, and $\hat{V}_{R,E}$ are defined as
\begin{align}
    \hat{Q}_{R,R}(s_t,a_t) &= r_t + \gamma \mathbb{E}_{s_{t+1}\sim P(\cdot|s_t,a_t)}[V_{R,\bar{\psi}_R}(s_{t+1})],\label{eq:pracrewq}\\
    \hat{V}_{R,R}(s_t) &= \mathbb{E}_{a_t\sim \pi_{T,\theta_T}}[ Q_{R,R,\phi_R}(s_t,a_t)],\label{eq:pracrewv}\\
    \hat{Q}_{R,E}(s_t,a_t) &= \gamma \mathbb{E}_{s_{t+1}\sim P(\cdot|s_t,a_t)}[V_{R,E,\bar{\psi}_E}(s_{t+1})],\label{eq:pracpureq}\\
    \hat{V}_{R,E}(s_t) &= \mathbb{E}_{a_t\sim \pi_{E,\theta_E}}[ Q_{R,E,\phi_E}(s_t,a_t) + \alpha_Q\log\pi_{E,\theta_E}(a_t|s_t)].\label{eq:pracpurev}
\end{align}
Here, $\bar{\psi}_R$ and $\bar{\psi}_E$ are the target network parameters and we consider $Q$-functions $Q_{R,R,\phi_{R,i}},~i=1,2$ for $Q_{R,R}$ and $Q_{R,E,\phi_{E,i}},~i=1,2$ for $Q_{R,E}$.

Other details are the same as those in the  implementation of vanilla MME. We summarize the propose algorithm in Algorithm \ref{alg:mme}.

\begin{algorithm}[h]
	\begin{algorithmic}
	    \IF{Disentangled}
		\STATE{Initialize $\theta_T$, $\theta_E$, $\psi_R$, $\bar{\psi}_R$, $\psi_E$, $\bar{\psi}_E$, $\phi_{R,i}$, $\phi_{E,i},~i=1,2$}
		\ELSE
		\STATE{Initialize $\theta$, $\psi$, $\bar{\psi}$, $\phi_i,~i=1,2$}
		\ENDIF
		\FOR{each iteration}
		\STATE{Sample a trajectory $\tau$ from the behaviour policy and store $\tau$ in the buffer $\mathcal{D}$}
		\FOR{each gradient step}
		\STATE{Sample a random mini-batch of size $M$ from $\mathcal{D}$}
		\IF{Disentangled}
		\STATE{Compute $\hat{J}_{\pi_T}(\theta_T)$, $\hat{L}_{Q_{R,R}}(\phi_{R,i})$, $\hat{L}_{V_{R,R}}(\psi_R)$ from the mini-batch}
		\STATE{Compute $\hat{J}_{\pi_E}(\theta_E)$, $\hat{L}_{Q_{R,E}}(\phi_{E,i})$, $\hat{L}_{V_{R,E}}(\psi_E)$ from the mini-batch}
		\STATE{$\theta_T\leftarrow \theta_T+\delta\nabla_{\theta_T}\hat{J}_{\pi_T}(\theta_T)$,\quad $\theta_E\leftarrow \theta_E+\delta\nabla_{\theta_E}\hat{J}_{\pi_E}(\theta_E)$}
		\STATE{$\psi_R\leftarrow \psi_R-\delta\nabla_{\psi_R}\hat{L}_{V_{R,R}}(\psi_R)$,\quad$\psi_E\leftarrow \psi_E-\delta\nabla_{\psi}\hat{L}_{V_{R,E}}(\psi_E)$}
		\STATE{$\phi_{R,i}\leftarrow \phi_{R,i}-\delta\nabla_{\phi_{R,i}}\hat{L}_{Q_{R,R}}(\phi_{R,i})$,\quad$\phi_{E,i}\leftarrow \phi_{E,i}-\delta\nabla_{\phi_{E,i}}\hat{L}_{Q_{R,E}}(\phi_{E,i}),~i=1,2$}
		\STATE{Update $\bar{\psi}_E$, $\bar{\psi}_R$ by EMA from $\psi_E$, $\psi_R$, respectively}
		\ELSE
		\STATE{Compute $\hat{J}_{\pi}(\theta)$, $\hat{L}_Q(\phi_i)$, $\hat{L}_V(\psi)$ from the mini-batch}
		\STATE{$\theta\leftarrow \theta+\delta\nabla_{\theta}\hat{J}_\pi(\theta)$}
		\STATE{$\psi\leftarrow \psi-\delta\nabla_{\psi}\hat{L}_V(\psi)$}
		\STATE{$\phi_i\leftarrow \phi_i-\delta\nabla_{\phi_i}\hat{L}_Q(\phi_i),~i=1,2$}
		\STATE{Update $\bar{\psi}$ by EMA from $\psi$}
		\ENDIF
		\ENDFOR
		\ENDFOR
	\end{algorithmic}
	\caption{(Disentangled) Max-Min Entropy RL}
	\label{alg:mme}
\end{algorithm}

\newpage
\section{Experimental Setup}
\label{sec:expsetup}

Here is the detailed setup for the overall experiments considered in this paper. Table \ref{table:parameter} shows the hyperparameter setup of MME and SAC. Basically, we follow the hyperparameter setup in the SAC paper \citep{haarnoja2018soft} for the parameters common to SAC and MME. Here, squashed Gaussian for the policy distribution means that the Gaussian policy is squashed into $[-1,1]$ by a tanh layer because we consider environments with bounded action space as proposed in \citep{haarnoja2018soft}. Table \ref{table:env} shows the detailed setup for all the considered environments in this paper. For the policy entropy coefficient $\alpha_{\pi}$ in Table \ref{table:env}, we considered the entropy coefficient value proposed in \citep{haarnoja2018soft, han2020diversity} for each task, and used the common $\alpha_{\pi}$ ($\alpha=\alpha_{\pi}$ for SAC), MME, and DE-MME. In addition, the value entropy coefficient $\alpha_Q$ in Table \ref{table:env} was chosen as the best performing value among $\alpha_Q\in\{0.1,0.2,0.5,1,2,5,10\}$ for each task. For more details, please refer to ablation studies on the entropy coefficients $\alpha_{\pi}$ and $\alpha_Q$ in Appendix \ref{sec:ablation}. For SparseMujoco tasks, the agent gets reward $1$ if it exceeds the $x$-axis threshold $\tau$ in Table \ref{table:env}, and for DelayedMujoco tasks, rewards are accumulated for $D$ time steps in Table \ref{table:env} and the agent gets the accumulated reward, as stated in Section \ref{subsec:compsparse}. We conducted all experiments in Section \ref{sec:exp} in environments where only CPUs are used without GPU. 
We used Mujoco and OpenAI Gym. We purchased a Mujoco license. OpenAI Gym is MIT license that allows use, copy, modification, merge, etc. So, it does cause any issue. 
The source code of the SAC baseline is open to public (MIT license), so using it does not have any issues. 
We did not use any data which contains personally identifiable information or offensive contents.

\begin{table}[!h]
	\centering
	\begin{tabular}{l|C{9em}C{9em}}
	\hline
	 & SAC & MME/DE-MME \\
	\hline
	Discount factor $\gamma$ & \multicolumn{2}{c}{$0.999$ for pure exploration / $0.99$ for rewarded setup} \\
	Learning rate $\delta$ & \multicolumn{2}{c}{$3\cdot 10^{-4}$} \\
	Episode length $N$ & \multicolumn{2}{c}{$1000$} \\
	Mini-batch size $M$ & \multicolumn{2}{c}{$256$} \\
	Replay buffer size & \multicolumn{2}{c}{$10^6$} \\
	Smoothing coefficient for EMA & \multicolumn{2}{c}{$0.005$} \\
	Optimizer & \multicolumn{2}{c}{Adam} \\
	Num. of hidden layers & \multicolumn{2}{c}{2} \\
	Size of hidden layers & \multicolumn{2}{c}{256} \\
	Activation layer & \multicolumn{2}{c}{ReLu} \\
	Output layer & \multicolumn{2}{c}{Linear} \\
	Policy distribution & \multicolumn{2}{c}{Squashed Gaussian distribution} \\
	\hline
	\end{tabular}
	\caption{Common hyperparamter setup}
	\label{table:parameter}
\end{table}

\begin{table}[!h]
	\centering
	\begin{adjustbox}{width=1\textwidth}
	\begin{tabular}{l|C{4.2em}C{4.7em}C{4.9em}C{4.7em}C{7em}C{4.8em}}
		\hline
		& State dim. & Action dim. & \makecell{$\alpha_{\pi}$ \\($\alpha$ for SAC)} & $\alpha_Q$(MME) &$\alpha_Q$(DE-MME) \\
		\hline
		ContinuousMaze & 2 & 2 & 1 & 0.5 & $\cdot$ \\
		HumanoidStandup-v1 & 376 & 17 & 1 & 2.0 & 0.1 \\
		Humanoid-v1 & 376 & 17 & 0.05 & 1.0 & 1.0 \\
		\hline
		 & State dim. & Action dim. & $\alpha_{\pi}$ & $\alpha_Q$(MME) & $\alpha_Q$(DE-MME) & Threshold $\tau$ \\
		\hline
		SparseHopper-v1 & 11 & 3 & 0.04 & 1.0 & 2.0 & 1.0 \\
		SparseHalfCheetah-v1 & 17 & 6 & 0.02 & 2.0 & 2.0 & 5.0 \\
		SparseWalker2d-v1 & 17 & 6 & 0.02 & 0.5 & 2.0 & 1.0 \\
		SparseAnt-v1 & 111 & 8 & 0.01 & 0.2 & 0.1 & 1.0 \\
		\hline
		 & State dim. & Action dim. & $\alpha_{\pi}$ & $\alpha_Q$(MME) & $\alpha_Q$(DE-MME) & Delay $D$ \\
		 \hline
		Del.Hopper-v1 & 11 & 3 & 0.2 & 1.0 & 2.0 & 20 \\
		Del.HalfCheetah-v1 & 17 & 6 & 0.2 & 2.0 & 2.0 & 20 \\
		Del.Walker2d-v1 & 17 & 6 & 0.2 & 0.5 & 2.0 & 20 \\
		Del.Ant-v1 & 111 & 8 & 0.2 & 0.2 & 0.1 & 20 \\
		\hline
	\end{tabular}
	\end{adjustbox}
	\caption{Detailed setup for environments}
	\label{table:env}
\end{table}

\newpage
\section{Additional Results on Performance Comparisons}
\label{sec:addperfcomp}

Here, we provide the max average return table of MME/DE-MME and SAC for experiments on sparse-rewarded tasks in Section \ref{subsec:mar}. In addition, we provide detailed explanation, implementation, and the result plots for performance comparison to recent exploration methods in Section \ref{subsec:compotherexp} and general RL algorithms in Section \ref{subsec:compothergeneral}.

\subsection{Max Average Return Results for Performance Comparison to SAC}
\label{subsec:mar}

Table \ref{table:marsac} shows the max average return performance of MME/DE-MME and SAC on sparse-reward (SparseMujoco and DelyaedMujoco) tasks in Section \ref{subsec:compsparse}. The value following the $\pm$ sign in the table means one standard deviation of the max average return, and the best result among the algorithms for each task is shown in bold. For all the considered tasks, both MME/DE-MME have the better max average return performance than SAC, and DE-MME has the best performance in most cases. Note that this result is  consistent with the results shown in Figs. \ref{fig:compsparsemujoco} and  \ref{fig:compdelayedmujoco}.

\begin{table*}[!h]
	\centering
		\begin{tabular}{L{7.1em}|C{7.5em}C{7.5em}C{7.5em}}
			\hline
			& MME & DE-MME & SAC\\
			\hline
			Sps.Hopper & {\bf902.50$\pm$4.36} & 893.30$\pm$6.72 & 823.70$\pm$215.35 \\
			Sps.HalfCheetah & 903.50$\pm$34.97 & {\bf924.90$\pm$39.57} & 386.90$\pm$404.70 \\
			Sps.Walker2d & 818.00$\pm$208.60 & {\bf886.60$\pm$25.77} & 273.30$\pm$417.51 \\
			Sps.Ant  & 953.70$\pm$28.39 & {\bf973.60$\pm$12.55} & 963.80$\pm$42.51\\
			Del. Hopper & {\bf3421.32$\pm$88.29} & {\bf3435.28$\pm$39.55} & 2175.31$\pm$1358.39\\
			Del. HalfCheetah & 7299.28$\pm$1562.19 & {\bf8451.20$\pm$1375.27} & 3742.33$\pm$3064.55\\
			Del. Walker2d & 5148.58$\pm$193.78 & {\bf5274.89$\pm$186.35} & 3220.92$\pm$1107.91 \\
			Del. Ant & 4664.04$\pm$836.37 & {\bf4851.64$\pm$830.88} & 3248.43$\pm$1454.48 \\
			\hline
		\end{tabular}
	\caption{Max average return of MME/DE-MME and SAC}
	\label{table:marsac}
	%\vspace{3em}
\end{table*}

\subsection{Comparison to Recent Exploration Methods on Sparse-Rewarded Tasks}
\label{subsec:compotherexp}

In Section \ref{subsec:compsparse}, we compared the performance of MME/DE-MME with various recent exploration methods on the considered sparse-rewarded Mujoco tasks. For comparison, we considered two types of exploration methods: action-based exploration that modifies the policy distribution itself to enhance exploration, and state-based exploration that finds rare states for better exploration. We first describe the detailed explanation and implementation for the considered recent exploration methods.

For action-based exploration methods, we considered 1) SAC with divergence regularization \citep{hong2018diversity} (SAC-Div), which adds a single diversity term $\alpha_d D(\pi||q)$ to the SAC objective for some divergence $D$ between the policy $\pi$ and the sample action distribution $q$ to choose actions away from the actions in the buffer and 2) diversity actor-critic \citep{han2020diversity} (DAC), which regularizes sample-aware entropy $\alpha\mathcal{H}(\beta\pi+(1-\beta)q),~\beta\in(0,1)$ instead of the policy entropy $\alpha\mathcal{H}(\pi)$ of SAC to enhance the policy entropy while the policy chooses actions to avoid the previously sampled actions in the buffer. 
For SAC-Div, we considered the KL divergence for $D$ and the adaptive scaling of $\alpha_d$ with scaling parameter $\delta_d = 0.2$, as suggested in \citep{hong2018diversity}. For DAC, we used the same $\alpha$ with the SAC baseline, and $\beta=0.5$ for SparseMujoco tasks and the adaptive  $\beta$ with control hyperparameter $c_{\mathrm{adapt}}=-2.0\mathrm{dim}(\mathcal{A})$ for DelayedMujoco tasks, as suggested in \citep{han2020diversity}.

For state-based exploration methods, we considered 1) random network distillation \citep{burda2018exploration} (RND), which uses the model prediction error $||f(s_{t+1})-f_t(s_{t+1})||^2$ as an intrinsic reward $r_{int,t}$ to search for rare states, where $f_s$ is a predictor network and $f_t$ is a randomly fixed target network, and 2) MaxEnt(State) \citep{hazan2019provably} which  maximizes the entropy of state mixture distribution  $\mathcal{H}(d^{\pi_{mix}})$ for maximizing $\mathcal{H}(d^\pi)$ to visit states uniformly, where $d^\pi$ is the state distribution induced by the policy $\pi$. 
MaxEnt(State) is 0riginally proposed for pure exploration, but we used the reward of MaxEnt(State) $-\log (d^{\pi_{mix}}(s) + c_s)$ as an intrinsic reward to search for rare states as the RND case for comparison on sparse-rewarded tasks, where $c_s$ is the smoothing constant in \citep{hazan2019provably}.
Then, the reward becomes $r_t=r_{ext,t}+c_{int}r_{int,t}$, where $r_{ext,t}$ is the external reward from the environment and $c_{int}>0$ is the intrinsic reward coefficient. For RND, we used MLP with 2 ReLU hidden layers of size $256$ and linear output layer of size $20$ for both predictor and target networks, and the intrinsic reward coefficient $c_{int}=5$ for sparse-rewarded Ant/HalfCheetah tasks and $c_{int}=10$ for sparse-rewarded Hopper/Walker tasks, properly chosen from $\{1,2,5,10\}$. For MaxEnt, we computed $d^{\pi_{mix}}$ based on projection/Kernel density estimation with Epanechnikov kernel using 100k previous states stored in the buffer, and used the intrinsic reward coefficient $c_{int}=0.01$ for sparse-rewarded Ant/HalfCheetah tasks and $c_{int}=0.02$ for sparse-rewarded Hopper/Walker tasks, properly chosen from $\{0.01,0.02,0.05,0.1\}$. For state-based exploration methods, we used the Gaussian policy with fixed standard deviation $\sigma=0.3$, and the other implementation setup was the same with the SAC baseline to make fair comparison between action-based and state-based exploration methods.

\begin{figure*}[!h]
	\centering
	\subfigure[SparseHopper-v1]{\includegraphics[width=0.24\textwidth]{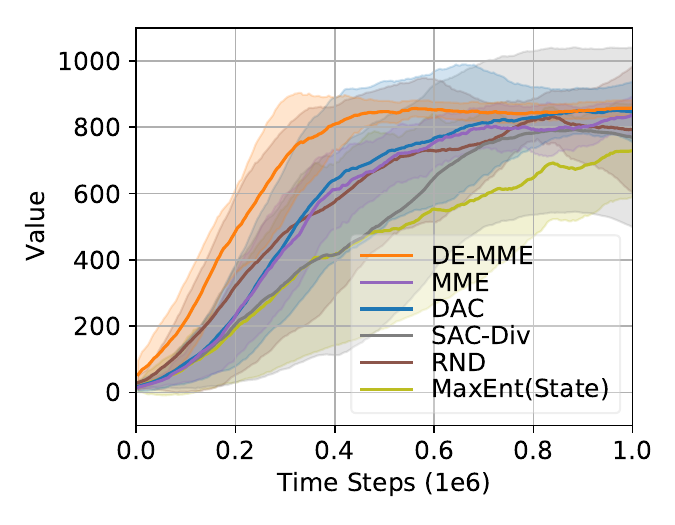}}
	\subfigure[SparseHalfCheetah-v1]{\includegraphics[width=0.24\textwidth]{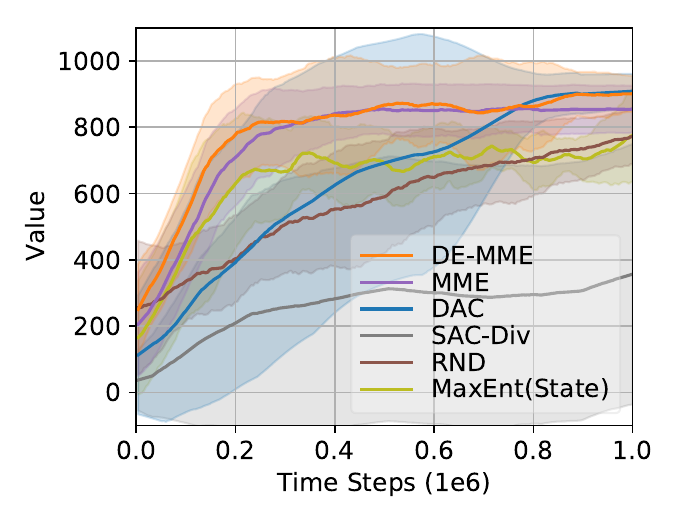}}
	\subfigure[SparseWalker2d-v1]{\includegraphics[width=0.24\textwidth]{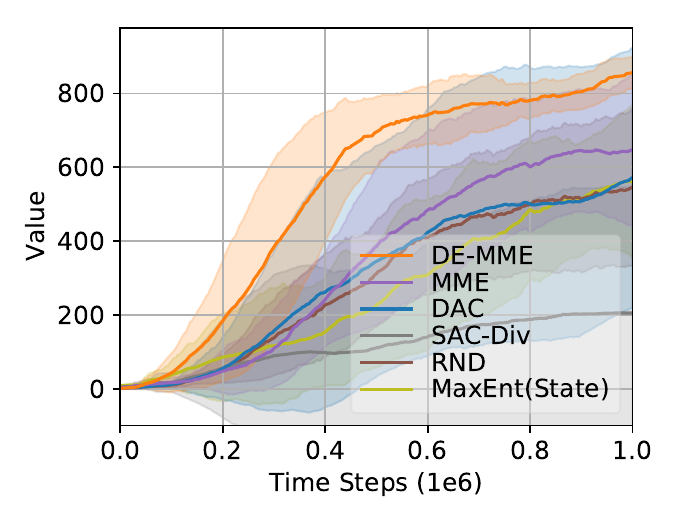}}
	\subfigure[SparseAnt-v1]{\includegraphics[width=0.24\textwidth]{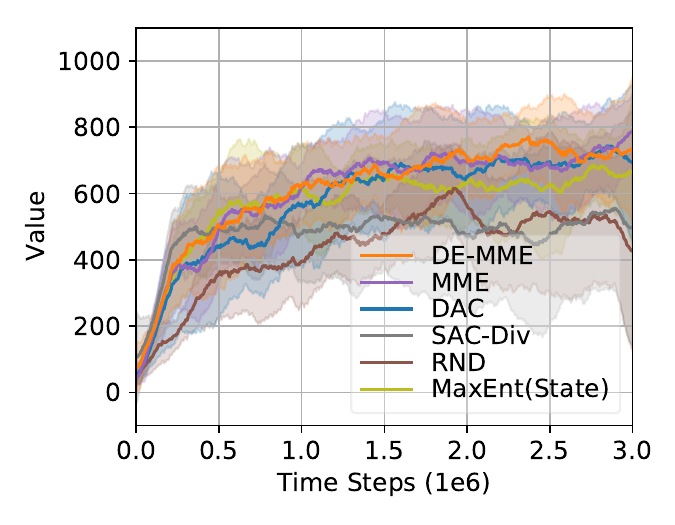}}
	\subfigure[Del.Hopper-v1]{\includegraphics[width=0.24\textwidth]{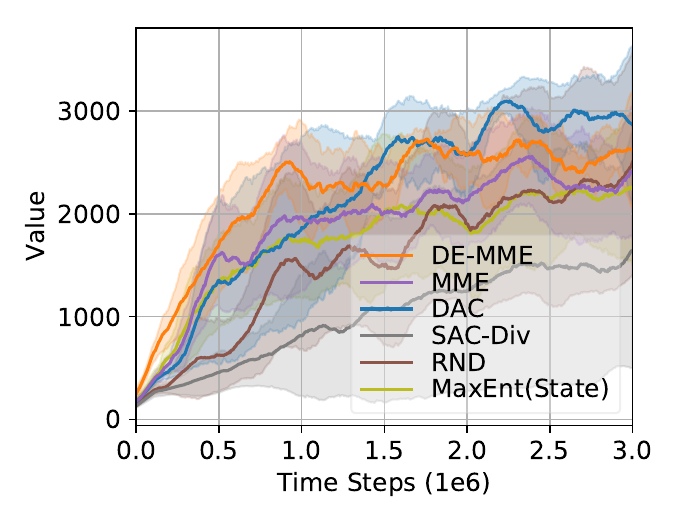}}
	\subfigure[Del.HalfCheetah-v1]{\includegraphics[width=0.24\textwidth]{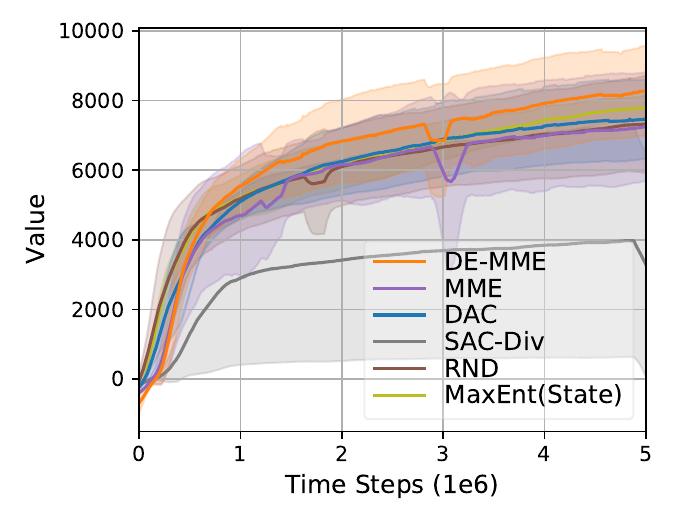}}
	\subfigure[Del.Walker2d-v1]{\includegraphics[width=0.24\textwidth]{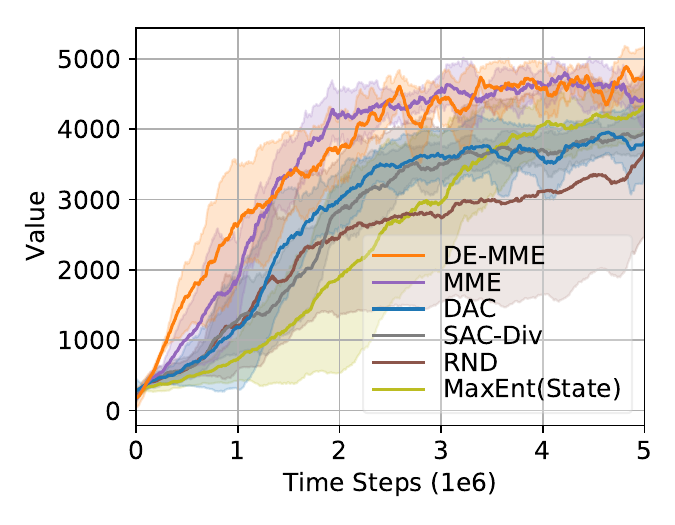}}
	\subfigure[Del.Ant-v1]{\includegraphics[width=0.24\textwidth]{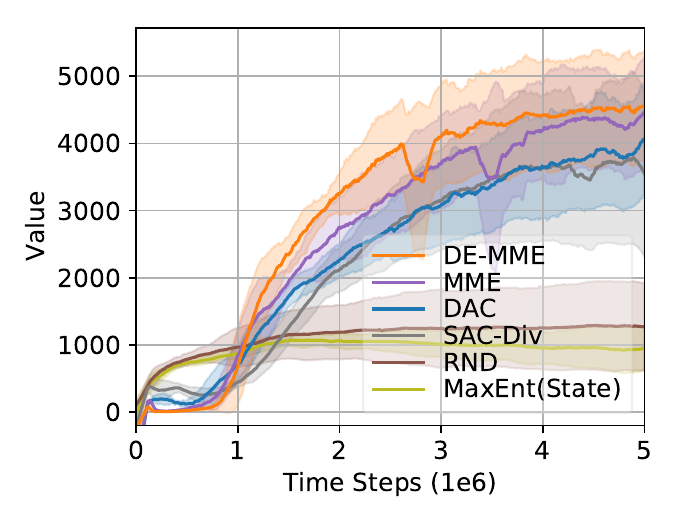}}
	\caption{Performance comparison to recent exploration methods on sparse-rewarded Mujoco tasks}
	\label{fig:compexp}
	\vspace{-3em}
\end{figure*}

Fig. \ref{fig:compexp} shows the corresponding average return performance on SparseMujoco tasks and DelayedMujoco tasks. It is seen that   proposed MME and DE-MME perform best in most of the considered tasks.

\newpage
\subsection{Comparison to Recent General RL Algorithms on Dense-Rewarded Tasks}
\label{subsec:compothergeneral}

In Section \ref{subsec:compsparse}, we  compared the performance of MME/DE-MME with various recent on-policy and off-policy RL algorithms on high action-dimensional dense-rewarded (original) Mujoco tasks (Humanoid, HumanoidStandup). The considered recent algorithms are as follows.
For on-policy RL, we considered trust region policy optimization (TRPO) \citep{nachum2017trust}, which  has a KL divergence constraint to guarantee the monotone improvement in policy gradient update, and proximal policy optimization (PPO) \citep{schulman2017proximal}, which efficiently restricts the amount of policy update by clipping the importance sampling ratio for stable learning. For both algorithms, we used implementations in the OpenAI baselines \citep{baselines}.

For off-policy RL, we considered two RL algorithms based on maximum entropy RL: soft actor-critic (SAC) \citep{haarnoja2018soft} explained in detail in Section \ref{subsec:sac}, and soft Q-learning (SQL) \citep{haarnoja2017rein}, which represents the policy with an energy-based model, where the energy function is the $Q$-function and uses Stein variational gradient descent  to learn the sampling network. For both algorithms, we used the implementations in authors' Github: \url{https://github.com/haarnoja/sac} for SAC and \url{https://github.com/haarnoja/softqlearning} for SQL.

\begin{figure*}[!h]
	\centering
	\subfigure[HumanoidStandup-v1]{\includegraphics[width=0.4\textwidth]{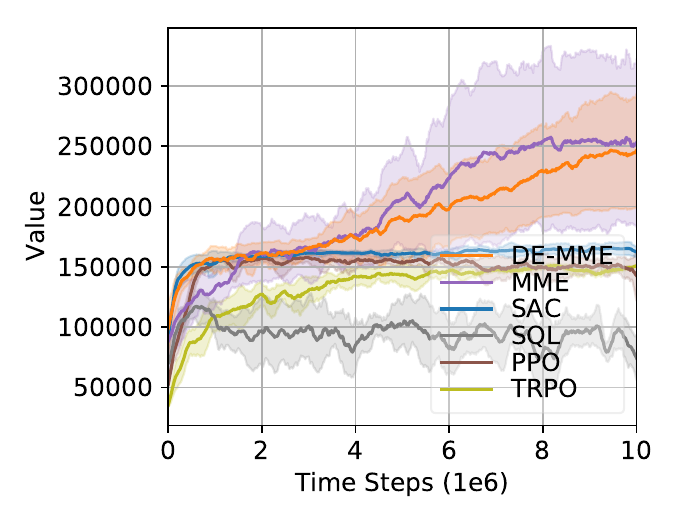}}
	\subfigure[Humanoid-v1]{\includegraphics[width=0.4\textwidth]{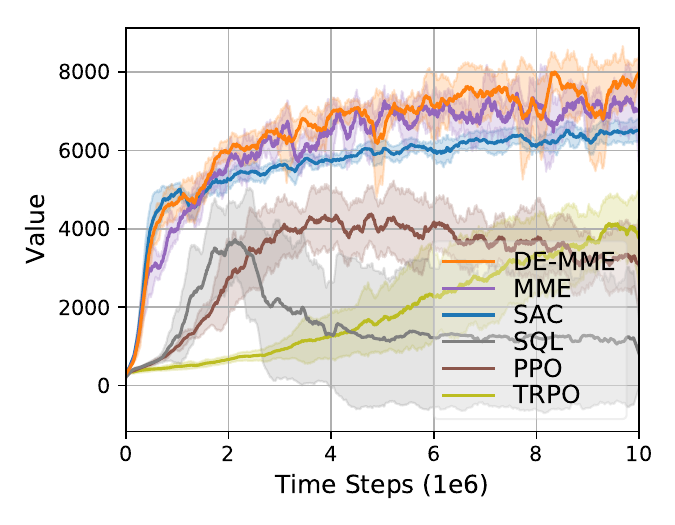}}
	\caption{Performance comparison to general RL algorithms on dense-rewarded Mujoco tasks}	\label{fig:compgeneral}
\end{figure*}

Fig. \ref{fig:compgeneral} shows the corresponding average return performance on Humanoid and HumanoidStandtup tasks. It is seen that  proposed MME and DE-MME yield superior performance to other RL algorithms for both tasks. Hence, we can observe that the proposed method is superior not only in sparse-reward environments but also in difficult dense-reward environments.

\newpage
\section{Ablation Studies}
\label{sec:ablation}

In this section of Appendix, we provide detailed ablation studies on the DelayedHalfCheetah task. For the hyperparameters not mentioned separately in each study, we used the hyperparameter setup in Table \ref{table:env} by default. 

\begin{figure}[!h]
    \centering
    \subfigure[Number of state visits]{\includegraphics[width=0.4\textwidth]{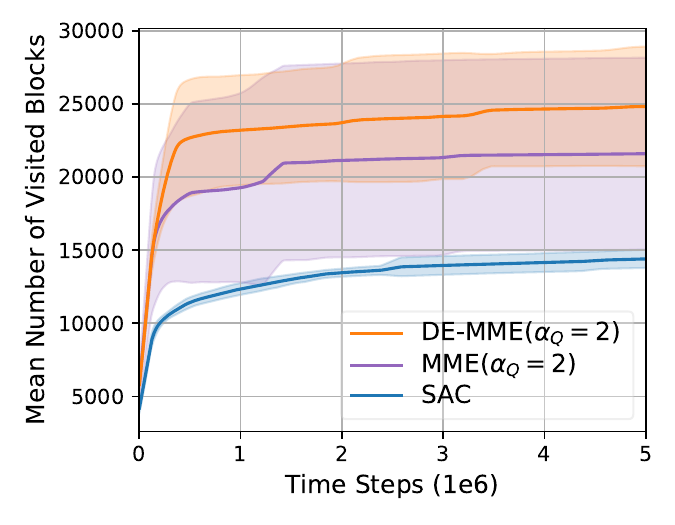}\label{fig:ablvisit}}
    \subfigure[Performance]{\includegraphics[width=0.4\textwidth]{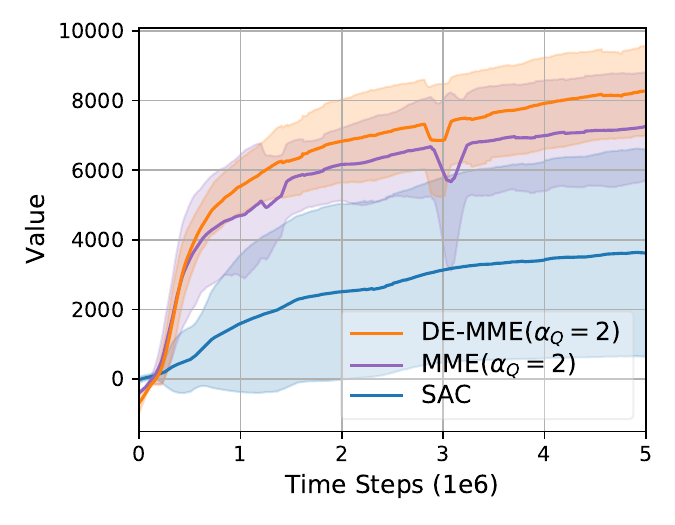}\label{fig:ablperf}}
    \subfigure[Performance for different $\alpha_Q$]{\includegraphics[width=0.4\textwidth]{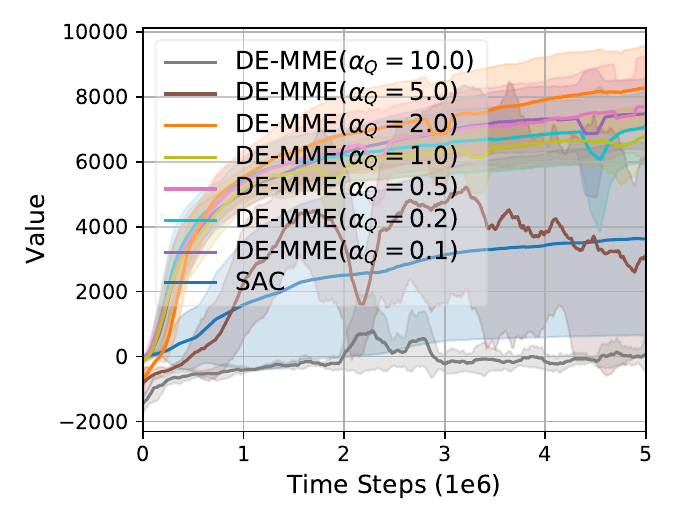}\label{fig:ablaq}}
    \subfigure[Performance for different $\alpha_\pi$]{\includegraphics[width=0.4\textwidth]{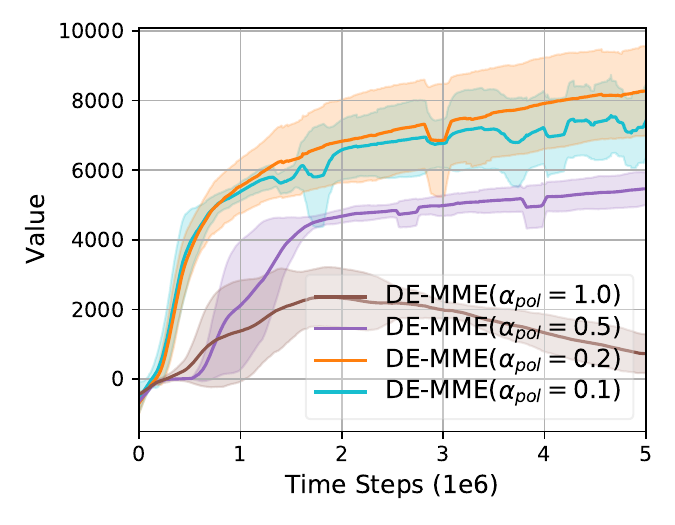}\label{fig:ablapol}}
    \caption{Ablation studies on DelayedHalfCheetah task}
    \label{fig:ablation}
\end{figure}

First, Figs. \ref{fig:ablation}\subref{fig:ablvisit} and \ref{fig:ablation}\subref{fig:ablperf} show the mean number of state visitation  and the corresponding performance, respectively, for   MME/DE-MME with $\alpha_Q=2.0$ and SAC.  Then, in order to see how $\alpha_Q$ and $\alpha_{\pi}$ affect the performance of DE-MME, we performed the same experiment with varying $\alpha_Q$ and $\alpha_\pi$, and the results are shown in 
 Figs. \ref{fig:ablation}\subref{fig:ablaq} and \ref{fig:ablation}\subref{fig:ablapol}.

{\bf MME vs. DE-MME:} Fig. \ref{fig:ablation}\subref{fig:ablvisit} shows the average number of quantized state visitation of MME/DE-MME and SAC. 
As in the pure exploration case, MME/DE-MME  visit much more states than SAC, which shows  the superior exploration performance of MME/DE-MME to SAC even in environments with reward. 
As mentioned  in Section \ref{sec:related},  enhanced exploration can help the policy not fall into a local optimum and converge better, and lead better final performance in various sparse-rewarded tasks, as shown 
in Fig. \ref{fig:ablation}\subref{fig:ablperf}.  It is also seen that DE-MME visits more states and has better performance than MME with the same $\alpha_Q$.  This shows that the disentangling exploration from exploitation proposed in Section \ref{subsec:disentangle} is indeed helpful for better exploration and performance.

{\bf Value entropy coefficient $\alpha_Q$:} The value entropy coefficient $\alpha_Q$ is one of the most important hyperparameters for MME/DE-MME, which determines the rate of change in the $Q$-function as the policy entropy changes. Thus, it affects the exploration from the $Q$-function to visit states with low entropy, as seen in Fig. \ref{fig:meanvisitmme} in the main paper. 
In tasks with reward, $\alpha_Q$ controls the ratio between the impacts of the reward function and the policy entropy on the  $Q$-function in the policy updates \eqref{eq:mmerpolup} and \eqref{eq:polupact}.   If $\alpha_Q$ is too large, the impact of the policy entropy becomes too dominant compared that of the reward function, and this adversely affects the performance,  as seen in Fig. \ref{fig:ablation}\subref{fig:ablaq}. On the other hand, if $\alpha_Q$ is too small, the impact of the $Q$-function to visit states with low policy entropy  for better exploration is reduced, so the performance again deteriorates, as seen in Fig. \ref{fig:ablation}\subref{fig:ablaq}. As a result, there is a trade-off between the reward function and the policy entropy in the $Q$-function, and we found that $\alpha_Q=2.0$ worked best for the DelayedHalfCheetah task.

{\bf Policy entropy coefficient $\alpha_{\pi}$:} Whereas the value entropy coefficient $\alpha_Q$ controls the entropy impact inside the $Q$-function to visit states with low entropy, the policy entropy coefficient $\alpha_{\pi}$ controls the ratio of $Q$-function itself to the policy entropy term to actually increase the policy entropy for given state in the policy updates  \eqref{eq:mmerpolup} or \eqref{eq:polupact}. 
If $\alpha_{\pi}$ is too large, the policy dominantly increases its entropy rather than increasing the expectation of $Q$-function including the reward sum in the policy update, so the performance deteriorates, as seen in Fig. \ref{fig:ablation}\subref{fig:ablapol}. If $\alpha_{\pi}$ is too small, on the other hand, then the policy is updated almost exclusively by the $Q$-function, so it is difficult to increase its  entropy and results in poor performance, as seen in Fig. \ref{fig:ablation}\subref{fig:ablapol}. 
We observed that $\alpha_{\pi}=0.2$ worked best for the DelayedHalfCheetah task. From the ablation studies, we conclude that it is important to choose appropriate $\alpha_Q$ and $\alpha_{\pi}$ to control the balance among the reward, the $Q$-function and the policy entropy.

\section{Answers to the Checklist}

\subsection{Limitations of Our Works}
\label{subsec:limitation}

As explained in Section \ref{sec:motivation}, SAC theoretically guarantees the optimal convergence for maximum entropy RL framework in finite MDP setup, but it shows  the saturation problem in practical  situation with function approximation and sample-based on learning when it learns the pure exploration task. 
In order to overcome this limitation, we proposed Max-Min entropy (MME) RL, which learns $Q$-function to explore states with low entropy and breaks the feedback loop of maximum entropy SAC that causes saturation. 

Although the proposed MME framework shows excellent performance compared to previous maximum entropy algorithms, we do not have proof of convergence for MME at this point even for finite MDPs. This theoretical work remains as a future work. Another issue is the complexity of MME. The complexity of DE-MME is larger than SAC since it has more parameters to learn than SAC. As a result, learning time can increase slightly compared to SAC. However, most RL algorithms consider sample complexity  most important, and DE-MME has superior performance to MME or SAC. Thus, we think that the increased complexity of DE-MME is well justified due to its superior performance.

\end{document}

%% file: macros.tex
\def\psfancypar#1#2{\begingroup\def\par{\endgraf\endgroup\lineskiplimit=0pt}
               \setbox2=\hbox{\large\sc #2}
               \newdimen\tmpht \tmpht \ht2 \advance\tmpht by \baselineskip
               \font\hhuge=Times-Bold at \tmpht
               \setbox1=\hbox{{\hhuge #1}}
               \count7=\tmpht \count8=\ht1
               \divide\count8 by 1000 \divide\count7 by \count8 
               \tmpht=.001\tmpht\multiply\tmpht by \count7 
               \font\hhuge=Times-Bold at \tmpht
               \setbox1=\hbox{{\hhuge #1}}
               \noindent
                \hangindent1.05\wd1
               \hangafter=-2 {\hskip-\hangindent
               \lower1\ht1\hbox{\raise1.0\ht2\copy1}%
                \kern-0\wd1}\copy2\lineskiplimit=-1000pt}

\newcommand{\E}{\mbox{{\rm E}}}

\def\boxit#1{\vbox{\hrule\hbox{\vrule\kern3pt
        \vbox{\kern3pt#1\kern3pt}\kern3pt\vrule}\hrule}}

\def\reals{ { {\rm  I \kern-0.15em R }  } }
\def\complex{ {\,{{\rm C} \kern-0.50em \raise0.20ex {  |}}\, }}

\def\Rbf{{\bf R}}

\def\Hc{{\cal H}}

\def\be{\vskip .3cm \begin{equation}}
\def\ee{\end{equation} \vskip .4cm \noindent}
\newcommand{\R}{\mbox{$\hat {\bf R}_{N}$}}

\def\Rxx{\Rbf_{\ssstyle X\kern-.1em X}}

\let\ssstyle=\scriptscriptstyle

\def\Kout{\setbox1=\hbox{\Huge\bf K}\hbox to
1.05\wd1{\hspace{.05\wd1}
}}
\def\Sout{\setbox1=\hbox{\Huge\bf S}\hbox to 1.05\wd1{\hspace{.05\wd1}
}}

%% file: labelfig.tex
  \ifx\LabelFigloaded\MYundefined\relax
  \else
   \fi

  \def\LabelFigloaded{\relax}% now loaded

  \chardef\LabelFigCatAt\the\catcode`\@
  \catcode`\@=11

 \let\LabelFigwlog@ld\wlog
 \def\wlog#1{\relax}

 \ifx\\\MYundefined@
    \let\\\relax
 \fi

  \def\ms@g{\immediate\write16}

 \def\N@wif{\csname newif\endcsname }
 \def\Temp@ {\N@wif\ifIN@}
 \ifx\INN@\MYundefined@
    \else \let\Temp@\relax
 \fi
 \Temp@

  \def\IN@{\expandafter\INN@\expandafter}
  \long\def\INN@0#1@#2@{\long\def\NI@##1#1##2##3\ENDNI@
    {\ifx\m@rker##2\IN@false\else\IN@true\fi}%
     \expandafter\NI@#2@@#1\m@rker\ENDNI@}
  \def\m@rker{\m@@rker}
 
  \newtoks\Initialtoks@  \newtoks\Terminaltoks@
  \def\SPLIT@{\expandafter\SPLITT@\expandafter}
  \def\SPLITT@0#1@#2@{\def\TTILPS@##1#1##2@{%
     \Initialtoks@{##1}\Terminaltoks@{##2}}\expandafter\TTILPS@#2@}

 \def\Shifted@@#1#2#3{\setbox0=\hbox{#3}%
   \raise -\dp0\vbox {\kern-#2%
       \hbox {\kern#1\unhbox0\kern-#1}%
           \kern#2}}

 \newcount\gridcount
 \newbox\auxGridbox@ \newbox\hGridbox@ \newbox\vGridbox@
 \newbox\Labelbox@ \newbox\auxLabelbox@
 \newbox\Coordinatebox@
 \newtoks\Labeltoks@
 \newdimen\Wdd@ \newdimen\Htt@
 \newdimen\Wddd@ \newdimen\Httt@
 
 \def\Wr@{\immediate\write16}

 \newdimen\GL@wd%% grid-line width
 \def\GridLineWidth#1{\GL@wd=#1}

 \def\gobble#1{}
 \def\EdgeErr@{\Wr@{}%
      \Wr@{\string\Edges\space argument
      1, 10, 100 or 1000 please\string!}%
      }

 \newcount\Edgect@

 \def\Sweepup#1\endSweepup{}

 \def\SetEdges@{%
    \edef\Zr@@s{\expandafter\gobble\number\Edgect@\empty}%
        \count255=0\Zr@@s\relax
        \ifnum\count255=\z@\else\EdgeErr@\show\tailtest\fi
        \count255=1\Zr@@s\relax%\showthe\count255
        \ifnum\count255=\Edgect@\relax\else\EdgeErr@\show\leadtest\fi
    \EdgGl@b\edef\Zr@s{\expandafter\gobble\Zr@@s\empty}%\show\Zr@s
    \ifnum\Edgect@>\@ne\relax\EdgGl@b\let\L@Dc\empty
        \else\EdgGl@b\edef\L@Dc{\string.}\fi
    \ifnum\Edgect@>\@ne\relax
        \EdgGl@b\edef\Edgescale@##1{\divide##1 by \Edgect@}%
        \else\EdgGl@b\edef\Edgescale@##1{}\fi
    }

 \def\Edges#1{\Edgect@=#1\relax
     \let\EdgGl@b\global \SetEdges@}

 \Edges{1}%% default

 \def\hhrule{\hrule height \GL@wd\vskip-.\GL@wd}

 \def\hRule@{%
   \advance\gridcount -2%
   \vfil\hhrule\vfil
   \llap{\smash{\raise -2.5pt
     \hbox{\L@Dc\number\gridcount\Zr@s\kern2pt}}}%
   \hhrule
   }

\def\vvrule{\vrule width \GL@wd \kern-\GL@wd}

 \def\vRule@{\advance\gridcount 2%
   \hfil\vvrule\hfil
   \setbox\auxGridbox@=\vbox to 0pt
      {\vskip \Htt@\vskip 2pt
        \hbox to 0pt{\hss\L@Dc\number\gridcount\Zr@s\hss}\vss}%
      \wd\auxGridbox@=0pt \box\auxGridbox@
   \vvrule
   }

 \def\PlaceGrid@@{\gridcount=10 
  \setbox\hGridbox@=\hbox{%
        \hbox{%
             \hskip-.4pt\vrule
             \vbox to \Htt@{%
               \offinterlineskip\parindent=\z@\relax
               \hbox to \Wdd@{\hfil}
               \hRule@\hRule@\hRule@\hRule@
               \vfil\hhrule\vfil}%
             \vrule\hskip-.4pt}
    }%
  \gridcount=0%
  \setbox\vGridbox@=\hbox{%
      \vbox{\offinterlineskip\parindent=0pt\hsize=0pt
         \vskip-.4pt\hrule%
         \hbox to \Wdd@{%
                 \vtop to \Htt@{\vfil}%
                 \vRule@\vRule@\vRule@\vRule@
                 \hfil\vvrule\hfil}%
         \hrule\vskip-.4pt}}%
  \wd\hGridbox@=0pt\ht\hGridbox@=0pt
  \wd\vGridbox@=0pt\ht\vGridbox@=0pt
  \hbox{\box\hGridbox@\box\vGridbox@}%
  }

 \def\LabelsGlobal{\def\LabGl@b{\global}}
 \def\LabelsLocal{\def\LabGl@b{}}
 \LabelsGlobal %% default

 \def\SetLabels#1\endSetLabels{%
   \LabGl@b\Labeltoks@={#1()\\}%
   }

 \LabGl@b\Labeltoks@={()\\}

 \def\ShowGrid{\LabGl@b\let\PlaceGrid@\PlaceGrid@@}
 \def\HideGrid{\LabGl@b\let\PlaceGrid@\relax}
 \def\Grids{\ShowGrid\LabGl@b\let\GridSwitch@\ShowGrid}
 \def\noGrids{\HideGrid\LabGl@b\let\GridSwitch@\HideGrid}

 \noGrids

 \def\bAdjust@@{%
     \setbox\auxLabelbox@=\hbox{\raise \dp\auxLabelbox@
            \box\auxLabelbox@}}
 \def\bAdjust@{\let\vAdjust@\bAdjust@@}

 \def\eAdjust@@{\dimen0=-.5\ht\auxLabelbox@
     \advance\dimen0 by .5\dp\auxLabelbox@
     \setbox\auxLabelbox@=
            \hbox{\raise\dimen0\box\auxLabelbox@}}
 \def\eAdjust@{\let\vAdjust@\eAdjust@@}

 \def\tAdjust@@{%
     \setbox\auxLabelbox@=\hbox{\raise-\ht\auxLabelbox@
            \box\auxLabelbox@}}
 \def\tAdjust@{\let\vAdjust@\tAdjust@@}

 \let\vAdjust@\relax

 \def\lAdjust@{\let\hAdjust@\rlap}
 \def\rAdjust@{\let\hAdjust@\llap}

 \let\hAdjust@\relax\let\vAdjust@\relax

 \def\FetchLabel@#1(#2)#3\\{%
     \IN@0#2@@\ifIN@
        \setbox0=\hbox{\ignorespaces#1#3\unskip}%
        \ifdim\wd0>0pt
           \ms@g{}%
           \ms@g{ !!! Bad label(s)? !!!}%
           \message{ #1(#2)#3}%
        \fi
        \def\LabelMole@##1\endFetchLabel@{%
            \IN@0()\\@##1@%
            \ifIN@\def\Temp@{\FetchLabel@##1\endFetchLabel@}%
            \else\def\Temp@{}%
            \fi
            \Temp@
           }%
     \else
       \ignorespaces#1\unskip
       \setbox\auxLabelbox@=%
         \hbox to 0pt{\hss\ignorespaces\hAdjust@
          {\ignorespaces#3\unskip}\hss}%
       \vAdjust@
       \let\hAdjust@\relax\let\vAdjust@\relax
       \AugmentLabelBox@@{#2}%
       \ht\Labelbox@=0pt\dp\Labelbox@=0pt
       \let\LabelMole@\FetchLabel@%
     \fi\LabelMole@}

 \newtoks\XYSep@ %\XYSep@{*}
 \def\SetXYSeparator#1{%
     \IN@0#1@@\ifIN@\XYSep@{*}%
     \else
     \XYSep@{#1}%
     \fi
     }

 \SetXYSeparator*

 \def\AugmentLabelBox@@#1{%
     \IN@0\the\XYSep@ @#1@\ifIN@
       \SPLIT@0\the\XYSep@ @#1@%
       \setbox\Labelbox@=\hbox to 0pt{%
         \unhbox\Labelbox@
         \Shifted@@{\the\Initialtoks@\Wddd@}%
         {\the\Terminaltoks@\Httt@}%
         {\box\auxLabelbox@}}%
     \else
         \ms@g{}%
         \ms@g{ !!! Bad insertion point. !!!}%
         \message{ (#1\ this point was rejected.)}%
     \fi
    }

 \def\FetchOption@#1[#2]#3\endFetchOption@{%
    \def\temp{#1}%\show\temp
    \ifx\temp\empty
       \Edgect@=#2\relax%\showthe\Edgect@
       \let\EdgGl@b\relax
       \SetEdges@%\def\Edgescale@##1{\divide##1 by \Edgect@\relax}%
       \Cleaner@#3%
    \fi}

 \def\Cleaner@#1[@]{\Labeltoks@{#1}}
     
 \def\PlaceLabels@@{\mathsurround=0pt%\bgroup
     \def\Cr@{\\}%
     \let\L\lAdjust@\let\R\rAdjust@
     \let\B\bAdjust@\let\E\eAdjust@\let\T\tAdjust@
     \expandafter\FetchOption@\the\Labeltoks@[@]\endFetchOption@
     \Wddd@=\Wdd@ \Edgescale@\Wddd@ %\showthe\Edgect@
     \Httt@=\Htt@ \Edgescale@\Httt@
     \expandafter\FetchLabel@\the\Labeltoks@\endFetchLabel@
     \box\Labelbox@%\egroup
     }%

 \let \PlaceLabels@\PlaceLabels@@

 \def\AffixLabels#1{\setbox\Coordinatebox@=\hbox{#1}%
      \Wdd@=\wd\Coordinatebox@ \Htt@=\ht\Coordinatebox@
      \advance\Htt@ \dp\Coordinatebox@
      \hbox{\copy\Coordinatebox@\kern-\Wdd@ 
           \Shifted@@{0pt}{-\dp\Coordinatebox@}%
           {\PlaceLabels@\PlaceGrid@}%
           \kern\Wdd@}%
      \GridSwitch@ %% next grid hidden
      \LabGl@b\Labeltoks@{()\\}%
      }
 
   \let\wlog\LabelFigwlog@ld   %%restore logging
   \catcode`\@=\LabelFigCatAt  %%12 or 13

                                By

              Raymond S\'eroul <A18645@FRCCSC21.BITNET>
                                and 
              Laurent Siebenmann <lcs@topo.math.u-psud.fr>
    
              VERSIONS: July 1991, Oct 1991, Jan 1992, July 1992

INTRODUCTION

      This labelling package is intended for TeX users who
rely on non-TeX sources for for their graphics inserts.  It
provides means for adding TeX labels to such inserts with a
minimum of fuss. 

       For most labels, TeX users have in the past found it
reasonably convenient to rely on non-TeX sources. Typical
occasions when an inescapable need for TeX labels seemed to
arise are

 (a) when the graphics program lacks certain exotic or complex
mathematical symbols

 (b) when the very highest typographical quality is wanted for the
labels

 (c) when labels included with the graphics fail to print, 
 and you cannot figure out why (cf. boxedeps.doc).  The labels
 provided by labelfig.tex are 100% portable.

       Since this package first appeared, many users, who in the
past scarcely dreamed of using TeX labels, have come to use
nothing but.  So it is now appropriate to add

Intoxication Warning:  TeX labels may be addictive and expensive. 

     If you have a fast preview you may disagree, and even find
that this package provides an agreeable paste-up environment; see
extra applications at end.

     Note to publishers: It is possible and convenient to ultimately
export the TeX labels produced by labelfig.tex to become an integral
part of the EPS file. This is often desired by a publisher who typically
uses an "upmarket" graphics or page layout program, with which the
staff is skilled in perfecting figures.  See Appendix I for
a recipe.

     The authors are grateful to Patrick Ion of Math Reviews for
helpful comments and encouragement.

BASIC INSTRUCTIONS

    After reading in the macro file using

\input labelfig.tex
preview or proof your figure with a coordinate grid printed on
top, by typing the following:

    \ShowGrid  % shows grid  for next figure only
    \AffixLabels{<the graphics insertion>}

Here <the graphics insertion> is what you would type to insert
the graphics object alone without the grid.  This must provide
for the space around it. For example <the graphics insertion>
might well be \BoxedEPSF{MyFigure scaled 700} using the
boxedeps.tex macro package (from same source); this provides a
TeX box containing the encapsulated PostScript insert specified by
the file MyFigure. \AffixLabels{...} provides the grid (supposing
\ShowGrid is present) and later, once you have specified labels
using the grid, it will "tack on" the labels.

     The grid is a sort of (usually elongated) checkerboard of
ten rows and ten columns and its (internal) partitions are by
default numbered  .1, ... ,.9  both horizontally (X-coordinate
running left to right) and vertically (Y-coordinate running bottom
to top).  Thus the points enclosed by the grid correspond to the
points of the unit square in the cartesian "X-Y" plane, the lower
left corner corresponding to the origin (0,0).  By extrapolation,
the full page corresponds to a larger rectangle in the plane.

     These coordinates serve to position labels as follows.
Before the \AffixLabels{...} command type label specifications:

  \SetLabels
   (<X-coordinate>*<Y-coordinate>) <first label> \\
   .
   .
   .
   (<X-coordinate>*<Y-coordinate>)  <last label> \\
  \endSetLabels

Each row specifies one label and is terminated by \\.  In each
row, the position indicator comes first; it is written as a
standard cartesian point except that the X- and Y- coordinates
are separated by * rather than a comma because TeX allows a
comma as decimal point. There are no dimension units to specify
as the unit is the grid itself.

     By default, this cartesian point specifies where the middle
of the baseline of the label will be located.  However if you precede
the point by \L [or \R] the left [or right] edge of the baseline will
be located there. Similarly you may also precede the point by \T, \E,
or \B to vertically align the top equator or bottom of the label box
at the specified point.  This gives nine standard positions of
the label with respect to the insertion point --- corresponding to
the eight principle points of the compas and the center

                     \L\T     \T      \R\T

                     \L\E     \E      \R\E

                     \L\B     \B      \R\B

But this neglects the default "baseline" level of TeX,
giving potentially three more positions

                     \L    <no tag>   \R

For text, the baseline level is often the preferred. Its relation to
the others is variable. It will often coincide with the bottom level,
as happens for "X".  But it is often distinct, as for "g", in which
case you have in all 12 distinct positions rather than 9.

     It is convenient to think of this specification of label
position as attaching the label by a thumb-tack to the coordinate
grid. There are up to twelve positions of the thumb-tack on the
label, while the position of the thumb-tack on the coordinate grid is
arbitrary.  Normally, one choses the position of the thumb-tack on
the label to be the one that is the closest to the item being
labeled.  There are good reasons for this "rule of thumb":

   (a)  It facilitates correct positioning at first try.

   (b)  If the scale of the figure must be altered after labels
have been affixed, the labels have a good chance of remaining well
positioned.

   (c)  The visible grid need not extend beyond the "bounding box"
for the figure, because the best preferred position is always
(at least almost) within the bounding box .

The second reason is particularly important. Indeed it often
happens that scale has to be altered after labelling begins, in
order to either provide space for the labels, or to adjust
proportions between the labels and the figure.  (The size of labels
is unaffected by scaling.)

     Here is an artificial but self-contained test which uses
TeX rules to make a graphics object.

TEST

    Do not skip this!

\input labelfig
 \def\FrameIt#1{\hbox{\vrule$\vcenter {\hrule\kern3pt%
             \hbox {\kern3pt #1\kern3pt}%
               \kern3pt\hrule}$\relax\vrule}}

 \def\Caption#1#2{\FrameIt{%
       \vtop {\hsize=#1\relax \parindent=0pt
         \leftskip=0pt \rightskip=0pt plus15pt
         \parfillskip=0pt
         \lineskip=1pt\baselineskip=0pt
         #2}}}

 \def\FirstQuadrant{\hbox to 100pt{\vrule\vbox to 100pt{%
        \hbox to 100pt{\hfil}\vfil\hrule}\hss}}

  \SetLabels
    \R(.5*.2) $\zeta\,\cdot$\\
    (.9*-.10) $\xi$\\
    \R(-.03*.9) $\eta$\\
    \T(.5*.9) \Caption{70pt}{%
          \it The norm of
          $g(\xi+i\eta)$ is indicated on
          contours of this invisible surface.}\\
  \endSetLabels

  \AffixLabels{\FirstQuadrant}

  \end

  Note that the coordinates to use for labels are indicated on the
edges of the grid (when visible) corresponding to the conventional
x- and y- axes of the Cartesian plane. By default the grid is
1-by-1. However, by the command \Edges{100}, you can change this
to 100-by-100 and many users find this alternative most
convenient. Place the command \Edges{...} in your style file (or
header) since its effect is is global. Other possible edge values
are 10 and 1000.

  If you use the command \Edges{...} at all, do so with care.  For
if you accidentally delete an \Edges{...} command your labels will
abruptly be badly misplaced and may logically but mysteriously
generate "dimension too big" errors under TeX and "off page" errors
under your driver.  

  You can dictate the edgescale for an individual figure by giving
the scale in brackets immediately after \SetLabels.  Thus, to
import into an article using say \Edge{100} a figure labelled using
another edgescale, say the original 1-by-1 default, you can use
\SetLabels[1]...\endSetLabels.

GETTING IT DOWN PAT

     Complicated labeling deserves the same respect as
complicated mathematics.  Do not expect it to come out perfect the
first time!  What is needed in either case is a mechanism to
repeatedly typeset troublesome pieces.

     One mechanism is always available.  One does complicated
labelling in a separate "test" file involving just the figure being
labelled;  a texpert will know how to \dump TeX's current state as
a temporary format that restarts rapidly at each retry.  Usually,
one then pastes the completed labelled figure back into the main
TeX file, but, of course, one can also \input it as an auxiliary
file.

     If you do not have a TeXpert at handy, here is a first
approximation to an efficient setup. By deletions reduce a copy
of your article to just a few lines before and after the figure.
Now label the figure, and finally, copy and paste the labelled
figure to the original article. Then copy the next figure to label
into this testbed and repeat. The TeXpert can improve the  speed
at which TeX starts up, by compiling a format specifically for
your article; just one caution: best NOT include in the format
ephemeral details of setup like \Set<mydriver>ArtSpecials (from
boxedeps.tex because this reads  figure dimensions which you may
change during your work session.

     An improved mechanism to repeatedly typeset troublesome
pieces is now available on the Macintosh; it is called LinoTeX;
see the same ftp sources.  It could be set up on many types
of computer.

     Before using labelfig.tex to attach labels to a graphics
object inserted using boxedeps.tex or BoxedArt.tex, make it a
firm rule to carefully adjust the bounding box using the trimming
commands of these packages, and also at least tentatively scale
and position the object. Beware of changing the grid inadvertently
after the labels have been positioned.  For example, correcting
the bounding box of a PostScript graphics object can foul up the
labels by changing the coordinate grid to which the labels are
attached. This is particularly true for the trimming  commands of
boxedeps.tex and BoxedArt.tex. However, as noted already, change
of scale is much less disruptive, and modest adjustments should be
well tolerated.

     Sometimes the labels protrude so far from the bounding box
of a figure that the figure has to be repositioned.  Best do this
by ad hoc spacing, say using \hglue and \vglue; altering the
bounding box would create a vicious circle.

     Remember that you are responsible for preventing labels
from overlapping. You are responsible for all label typography
including size and style. A label is really just about anything
that can be put in a TeX box. Note that spaces at the beginning
and end of labels will normally be suppressed; if you really want
them you must protect them with TeX braces.

     This package temporarily sets the \mathsurround parameter
of TeX to zero  while the labels are being affixed. This is done
because nonzero \mathsurround space would influence the position
of left and right aligned labels; then, when a texpert or printer
modifies mathsurround, diagram labeling might be disastrously
altered. There is a small price to pay involving labels that are
formatted as caption boxes including mathematics: you  may want or
need to specify an explicit mathsurround space within the caption
box; it will not influence anything outside.

     Those hostile to the use of * as separator between
the X and Y coordinates of label insertion points, are free to
impose another using \SetXYSeparator{<the new separator>}.  
Americans may prefer "," to "*" since they never use a 
comma as a decimal point; on the other hand, * may be more visible.

APPENDIX (I)  MERGING labelfig.tex LABELS INTO AN EPSF GRAPHICS OBJECT.

     As promised in the introduction, here is a recipe useful for
publishers. It works at least on Macintosh and at least for vectorized
graphics and Adobe type1 fonts.  (There is surely a similar recipe for
PCs under MSWindows.)

 (a)  Use boxedeps.tex utility to integrate the figure given by the eps
file, "x.eps" say, with a visible frame around it.  See
\ShowDisplacementBoxes command in boxedeps.tex.  To get precise results
automatically it is important to use the \Trim... commands of
boxedeps.tex making the "DisplacementBox" neatly fit the figure.

 (b)  Use the TeX printer driver and LaserWriter (versions >= 8.1.1) to
export to an EPSF the DVI page containing the integrated, labelled
figure. You now have an EPS file  "xx.eps"  that contains too much, and at
the wrong scale, and at wrong position.

 (c)  Convert the EPSF to an Adode Illustrator format EPSF using
the shareware utility called epsConvert by Sam Weiss
1993-- (currently $25).

 (d)  In Illustrator (or a compatible program), group the labels and the
"DisplacementBox"; copy them to the clipboard and paste them into "x.ps".
This step requires that all the label fonts be "visible to the Macintosh.

 (e)  Translate and scale the pasted group consisting of the labels plus
the "DisplacementBox" so as to make the "DisplacementBox" the bounding
box of (labelless) figure represented by "x.eps".  At this point the
labels will be correctly placed on the figure "x.eps".

 (f)  Ungroup and delete the "DisplacementBox".  The result is the
desired single EPS file, "x+.eps" say, It contains the original figure
plus its labels.  

     Using grouping and ungrouping appropriately in "x+.eps", a
publisher's staff can very efficiently improve label positions etc.

APPENDIX II)  SOME EXOTIC APPLICATIONS

     The grid of labelfig.tex is analogous to a light-table in
classical page makeup with wax or latex glue.  In principle, you
can use it to compose any page from its indivisible parts.  This
even has some of the artisanal charm of classical paste-up
provided you have a fast screen preview to make the process
"interactive".

     In practice labelfig.tex is a tool for nonstandard jobs.
Here are a few going beyond the labelling already discussed.

(I)  GRAPHICS INTEGRATION.

     This is accomplished by treating the imported graphics
objects as labels.  The underlying graphics object is then
typically an empty  \vbox to <dimension>{\vfill} in a TeX
\midinsert...\endinsert construction.  A label line
might be of the form

   (.1*.1) \\

The exact form of the special command varies from driver to
driver.  However, in the case of encapsulated PostScript graphics
(EPSF norm), by relying on boxedeps.tex, one can have the
following standard syntax (independant of driver  (see
boxedeps.doc for details.
  
  (.1*.1) \BoxedEPSF{MyFigure scaled <scale in mils>}\\

This may be slow since it requires TeX to read the PostScript
file to read bounding box using many complex macros.  So you
may want to try

  (.1*.1) \EPSFSpecial{MyFigure}{<scale in mils>}\\

which is fast and driver independant, but it squashes the
bounding box, normally to its lower left corner.

     Similarly for graphics of the Macintosh PICT norm ---
using BoxedArt.tex (same sources) in place of boxedeps.tex.

     This approach to integration is to be recommended when
one is assembling a composite graphics object.

 (II)  COMMUTATIVE DIAGRAM ENHANCEMENT

     Commutative diagrams or arrays of mathematical objects
connected by arrows of various sorts are common in mathematics.
The mathematical objects require the use of TeX.  Recently TeX
acquired a good collection of arrows of all slopes --- that of
LamSTeX --- plus pwerful macros to build the diagrams.

     However, even the LamSTeX collection is often
inadequate; it lacks for example double shafted arrows, dotted
arrows and curved arrows. Fortunately it is possible to produce
such arrows on an individual basis using sophisticated graphics
programs such as Illustrator and AldusFreehand (both serving
the EPSF norm) or using Metafont (with its public domain norm).
Since the creation of each new arrow is a work of love, you
probably want to limit the number of arrows by using LamSTeX
for most arrows. The 40K commutative diagram module of LamSTeX
has been adapted to work with AmSTeX and a copy may be posted
with LabelFig and related files. Unfortunately no one has yet
offered a version that works with Plain TeX or LaTeX.

       Suffice it here to say that when the exotic arrow has
been somehow imported into TeX, labelfig.tex treats it as a
label that one affixes to the commutative diagram.  Two other
steps will be treated in separate notes, namely the matter of
extracting the dimension specifications for the arrow and the
construction of the arrow --- for these steps are far from
unique and often depend intimately on your computer environment. 
Notes for the Macintosh-Textures-Illustrator combination are
found in the file ExoticArrows.doc.

 (III) NESTING 

Ingenuity pays off in exploiting labelfig.tex. One can
mix graphics and typography quite freely.  labelfig.tex is good
for freeform or overlapping arrangements, while boxedeps.tex (or
BoxedArt.tex) is best for regimented non-overlapping
arrangements --- and the two can be combined.

     The default behavior of labelfig.tex is not ideal 
for nesting objects, because to prevent trouble for beginners
the register for labels is globally cleared when \AffixLabels
concludes.  But there are switches available

      \LabelsGlobal      \LabelsLocal

which change this.  To understand this, extend the above test 
by something like:

 \LabelsLocal
 \SetLabels
    (.5*.5) AAA\\
 \endSetLabels

 {%%% Watch for influence of braces!!
 \SetLabels
    (.5*.5) ZZZ\\
 \endSetLabels
   \AffixLabels{\FirstQuadrant}
 }

   \AffixLabels{\FirstQuadrant}

     There are however potential pitfalls.  Neither
labelfig.tex nor boxedeps.tex has been tested under extreme
conditions. Problems may occur if their procedures are
indiscriminately nested. For boxedeps.tex (not labelfig.tex)
there is a precise cause for worry, namely many of its
variables are "global", which means that TeX braces will not
provide the protection one might expect.

COMMAND SUMMARY FOR labelfig.tex

  Here [...] means optional (one or zero)
       [...]* means any number of such constructs

  \SetLabels
    [[<P>](<X><Sep><Y>) <label> \\]*
  \endSetLabels
  \ShowGrid  % this makes the grid visible (once)
  \AffixLabels{<the figure>}

   --- <P> is tack position, one of eleven or empty
              order irrelevant

                   \L\T      \T      \R\T

                   \L\E      \E      \R\E

                     \L               \R

                   \L\B      \B      \R\B

   --- (<X><Sep><Y>) insertion point;
  <Sep> is separator, = * by default;
  \SetXYSeparator{<Sep>} changes it.
   <X> and <Y> are real numbers

  --- <label> a label to attach 

  --- <the figure> the figure to label 

  \GlobalLabels (default)     
  \LocalLabels  setting for nested constructs.

 \Grids makes ALL grids appear; \HideGrid then makes just next disappear.
 \noGrids returns to default.  The commands are always global.

 \GridLineWidth{<dimension>} adjusts width of grid lines. Default is very
small, to give "hairline" effect. If your grid lines are missing try
setting \GridLineWidth{1pt}.

 \Edges#1 globally changes the edge size of all grids to the numerical 
value #1, which must be 1, 10, 100, or 1000.  The default is 1.

VERSION HISTORY.
 --- Jan 1993: \Edges#1 and [??] option after \SetLabels
 --- July 1992: \Grids, \noGrids, \HideGrid;
       Gridlines become hairlines; \GridLineWidth{<dimension>}.
 --- Oct 1991, Jan 1992: \SetXYSeparator{<Sep>},  \LabelsGlobal,
       \LabelsLocal.
 --- July 1991: first release

Address for bugs and other feedback:

        Raymond S\'eroul
        IREM and Lab. de Typographie Informatise
        Univ. Rene Descartes
        Strasbourg

    Tel 33-88-41-63-45
    Email:  A18645@FRCCSC21.BITNET

        Laurent Siebenmann
        Mathematique, Bat. 425,
        Univ de Paris-Sud,
        91405-Orsay,
        France

    Tel 33-1-6941-7949; 
    Email: lcs@topo.math.u-psud.fr